\documentclass[10pt,twocolumn,letterpaper]{article}

\usepackage[pagenumbers]{wacv} %

\usepackage[accsupp]{axessibility}

\definecolor{wacvblue}{rgb}{0.21,0.49,0.74}
\usepackage[pagebackref,breaklinks,colorlinks,allcolors=wacvblue]{hyperref}
\usepackage{graphicx}
\usepackage{multirow}
\usepackage{amsmath}
\usepackage{amssymb}
\usepackage{booktabs}
\usepackage[capitalize,nameinlink,sort&compress]{cleveref}
\usepackage{caption}
\usepackage{subcaption}
\usepackage{adjustbox}
\usepackage{rotating}
\usepackage{algpseudocode}
\usepackage{algorithm}
\usepackage{xcolor}
\usepackage{soul}
\definecolor{revblue}{RGB}{204,229,255}
\sethlcolor{revblue}

\soulregister\cite7
\soulregister{\cite}{0}
\soulregister{\cref}{0}
\soulregister{\Cref}{0}
\soulregister{\ref}{0}
\soulregister{\textbf}{0}
\soulregister{\textit}{0}
\soulregister{\emph}{0}
\soulregister{\texttt}{0}
\def\wacvPaperID{1274} %
\def\confName{WACV}
\def\confYear{2027}

\title{VISAGE: Vertex-based Intrinsic Surface Appearance GEneration via Dual-Stream Diffusion for Arbitrary 3D Heads}

\author{
Jeremias Krau{\ss}$^{1,2,*}$ \quad
Pragati Jaiswal$^{2,*}$ \quad
Ren\'{e} Schuster$^{1,2}$ \quad
Didier Stricker$^{1,2}$\\
$^{1}$German Research Center for Artificial Intelligence (DFKI)\\
$^{2}$RPTU Kaiserslautern-Landau\\
{\tt\small \{jeremias\_samuel.krauss, pragati.jaiswal, rene.schuster, didier.stricker\}@dfki.de}
}

\begin{document}
\maketitle
{\let\thefootnote\relax\footnotetext{
$^{*}$Equal contribution \\
Project page: \url{https://dfki-av.github.io/VISAGE}
}}
\begin{abstract}
Recovering identity-faithful, physically based surface appearance for 3D human heads from sparse images remains challenging. Existing methods often rely on UV-based pipelines, introducing seam artefacts, or operate in image space, limiting surface consistency. We present VISAGE, \textbf{V}ertex-based \textbf{I}ntrinsic \textbf{S}urface \textbf{A}ppearance \textbf{GE}neration, a surface-based conditional diffusion framework that synthesises per-vertex albedo and roughness directly on arbitrary mesh topologies, without requiring consistent UV parameterisation at training time. VISAGE combines global spectral processing with local geodesic K-Nearest-Neighbour refinement, enforcing structural coherence while restoring detail beyond the truncated spectral basis. Multi-view RGB images condition the network via visibility-aware cross-attention, enabling occlusion-robust synthesis across the full head surface. Region-specific modules specialise in semantically demanding and distinct areas (eyes, mouth, skin and hair), while a tangent-space output projection recovers fine iris detail. We further introduce a scalable procedural augmentation pipeline that enriches raw geometry scans with physically plausible albedo, roughness and normal maps for training. The predicted attributes render directly as a single RGBA vertex-colour channel. VISAGE reduces FID by 42.7\% on NPHM and 22.5\% and 4.4\% on zero-shot FaceVerse and i3DMM, respectively, relative to the strongest baseline, with a 7.2M-parameter core network.
\end{abstract}
    
\section{Introduction}
\begin{figure}[tb]
    \centering
    \includegraphics[width=\linewidth]{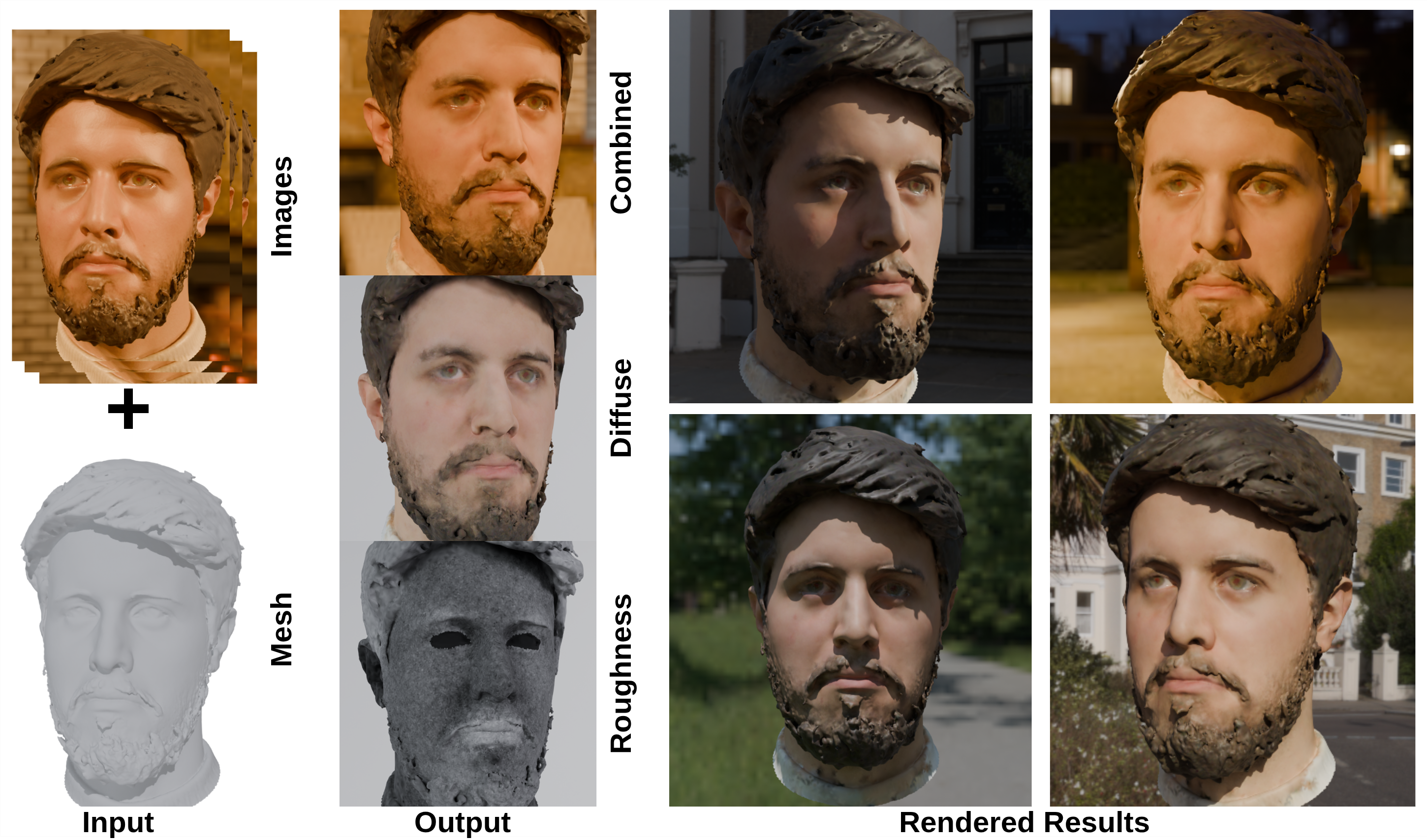}
    \caption{Given sparse 1-5 RGB images and an untextured head mesh (left), VISAGE predicts per-vertex albedo and roughness directly on arbitrary mesh topologies (center). The predicted intrinsic materials enable relighting under unseen illumination (right).} 
    \label{fig:Teaser}
\end{figure}

Digital human heads are central to telepresence, gaming, and AR/VR, where avatars are increasingly reconstructed from sparse imagery captured on consumer devices~\cite{cheng2024first}. Appearance captured under such conditions is entangled with scene illumination, so reproducing it directly restricts the avatar to that environment. Faithful rendering in a novel scene instead requires decomposed material attributes that disentangle surface reflectance from incident illumination, enabling relighting under arbitrary conditions with view-consistent appearance.

Traditional high-quality texturing relies on complex, costly capture stages~\cite{Borshukov2003RealisticHF,5473199,articleRelight}, while producing high-quality results, they are impractical for consumer applications. Learning-based methods that instead operate on unconstrained imagery avoid this cost, but often bake illumination into the predicted albedo or omit a decomposed material model altogether. Progress is further bottlenecked by the absence of large-scale datasets with ground-truth PBR head textures, detailed in \cref{sec:datagen}.
To address these challenges, we introduce VISAGE (\cref{fig:Teaser}), a surface-based diffusion framework for arbitrary head meshes. To overcome the lack of PBR ground-truth data for full 3D heads, we construct a scalable procedural augmentation pipeline in an off-the-shelf physically based rendering engine that enriches raw geometry scans with PBR materials, enabling large-scale training without costly light-stage capture. 
Operating directly on the mesh surface removes the need for consistent surface parameterisation, allowing the same network to operate on raw scans, remeshed geometry, and artist-defined topologies. Our dual-stream architecture combines global spectral diffusion for structural coherence with local geodesic K-Nearest-Neighbour (KNN) refinement for details. Conditioned via visibility-aware multi-view cross-attention on 1 to 5 unconstrained RGB images, VISAGE predicts view-consistent PBR-compatible per-vertex albedo and roughness maps. VISAGE also remains lightweight, with a 7.22M-parameter core generative network and competitive inference efficiency.
Our contributions are:
\begin{itemize}
\item \textbf{VISAGE:} A mesh-native surface diffusion model with two components specific to the conditional setting: heat-diffusion times conditioned on the noise level, and a parallel geodesic-KNN attention stream carrying the frequencies discarded by the truncated Laplace--Beltrami basis. Both are ablated in \cref{tab:ablation}. It predicts per-vertex albedo and roughness from sparse RGB views, requiring no consistent UV parameterisation for training; the output renders as a single RGBA vertex-colour channel.

\item \textbf{Visibility-aware multi-view conditioning:} A cross-attention mechanism that injects multi-view image features and intrinsic mesh descriptors into the diffusion process, combined with semantic region-specific modules and tangent-space refinement to specialise generation for demanding areas such as eyes, mouth, skin, and hair.

\item \textbf{Scalable procedural training data pipeline:}
A physics-inspired data generation pipeline that augments raw 3D head scans with plausible PBR material maps, enabling large-scale supervised training without expensive light-stage capture.

\end{itemize}

\section{Related Work}
\label{sec:related}
Diffusion-based methods for object texturing such as DreamFusion \cite{DreamFusion} and Fantasia3D \cite{Fantasia3D} have improved mesh texturing quality. To improve efficiency, projection-based pipelines including TEXTure \cite{TEXTure}, TexFusion \cite{texfusion}, Text2Tex \cite{Text2Tex}, Paint3D \cite{Paint3D}, MVDream \cite{MVDream}, and Hunyuan3D-2.0-Paint \cite{Hunyuan3d} generate textures through multi-view diffusion and UV-space aggregation. Although effective, these methods remain dependent on UV parameterisation and image-space consistency constraints.

To avoid UV artefacts, several recent methods operate directly on surface representations. UV3-TeD \cite{UV_FREE} employs intrinsic diffusion operators based on DiffusionNet \cite{DifNet}, Point-UV Diffusion \cite{PointDiffusion} combines point-based diffusion with UV refinement, and Texturify \cite{Texturify} introduces mesh-aware convolutions. These works demonstrate the potential of mesh-native learning and motivate our focus on surface-based diffusion for texture synthesis. UV3-TeD is the closest to our formulation but is unconditional: sampling depends only on geometric descriptors and the denoising timestep, with no mechanism for conditioning on photographs. It therefore generates plausible textures for a shape class rather than recovering subject-specific appearance. We instead extend DiffusionNet with multi-view cross-attention to form a surface-diffusion baseline. Moreover, UV3-TeD represents textures as point clouds requiring a dedicated renderer, whereas we bind appearance directly to vertices for native rendering.
\subsection{Face and Head Texturing and Generation}
Classical face modelling approaches such as 3D Morphable Models \cite{blanz2023morphable,5279762, Flame} learn low-dimensional parametric representations of facial geometry and appearance. While effective for reconstruction and control, they are limited in their ability to represent high-frequency appearance details. Neural extensions such as NPHM \cite{NPHM} and MonoNPHM \cite{MonoNPHM} increase geometric flexibility but focus primarily on reconstruction rather than texture synthesis.

A complementary line of work estimates facial reflectance and material properties through inverse rendering and differentiable shading \cite{Galanakis_2025_WACV,lattas2021avatarme++,InverseFaceNet,Dib_2024_CVPR}. Relightify~\cite{relightify} recovers relightable facial reflectance from a single image using a diffusion prior in UV space, but operates on a fixed template topology and has no public implementation at the time of writing. Progress in this direction is constrained by data. FFHQ-UV-Intrinsics~\cite{Dib_2024_CVPR} provides decomposed intrinsic maps at scale, but covers the face region only. MultiFace~\cite{wuu2022multiface} offers real multi-view capture with tracked low-resolution geometry, but under fixed illumination and without any reflectance decomposition. Neither provides full-head geometry paired with a decomposed albedo--roughness pair, which motivates the procedural pipeline of \cref{sec:datagen}.

Recent 3D-aware GANs and diffusion models --- EG3D~\cite{chan2022efficient},
PanoHead~\cite{an2023panohead}, FaceLift~\cite{lyu2024facelift},
Rodin~\cite{Rodin,RodinHD} and UltrAvatar~\cite{UltraAvatar} --- achieve
strong realism through volumetric, tri-plane or template-based
representations, but tie generated appearance to fixed parameterisations or
UV space.

\textbf{Positioning.} We build on DiffusionNet's intrinsic operators~\cite{DifNet}, EDM preconditioning~\cite{EDM} and surface-native texture diffusion~\cite{UV_FREE}; Our contribution is the adaptation to conditional, subject-specific head appearance via noise-conditioned spectral propagation and a parallel stream compensating for basis truncation.

\section{Method}
\begin{figure*}[tbh]
    \centering
    \includegraphics[width=\linewidth]{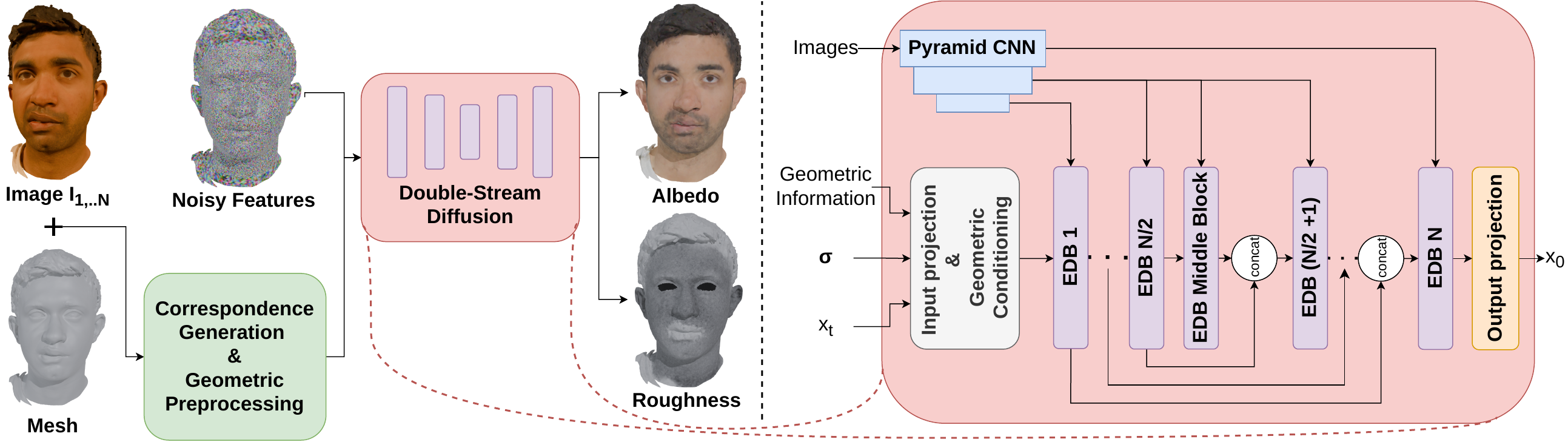}
    \caption{\textbf{Overview of VISAGE pipeline.} Noisy vertex features $x_t$ are concatenated with geometric descriptors and conditioned on multi-view image features before entering the dual-stream diffusion network. The network combines global spectral propagation with local geodesic refinement to predict the denoised representation $x_0$, which is projected to per-vertex albedo and roughness.}
    \label{fig:pipeline}
\end{figure*}

Given an untextured head mesh $\mathcal{M} = (\mathcal{V}, \mathcal{F})$ with arbitrary topology and a set of $N$ unconstrained RGB images $\mathcal{I} = \{I_i\}_{i=1}^N$ (with $1 \leq N \leq 5$ and $N=3$ as the default configuration), our goal is to synthesise physically plausible per-vertex albedo $\mathbf{a} \in \mathbb{R}^{|\mathcal{V}| \times 3}$, roughness $\mathbf{r} \in \mathbb{R}^{|\mathcal{V}|}$. Because albedo and roughness are stored as vertex attributes, they render without conversion: together they occupy exactly one standard RGBA vertex-colour channel, supported natively by common engines (Blender \cite{Blender} colour attributes, Unreal's \texttt{VertexColor} material node). Where a texture-mapped asset is required, the same attributes bake to a UV map without retraining.

We formulate texture synthesis as a conditional diffusion process directly on the mesh surface using intrinsic diffusion operators derived from DiffusionNet~\cite{DifNet}. Given vertex features $\mathbf{x}$, information is propagated over the manifold via the Laplace--Beltrami operator $\Delta$, enabling topology-agnostic feature exchange without UV parameterisation. 
While intrinsic diffusion provides strong global coherence, it inherently suppresses detail. To address this limitation, VISAGE employs a Dual-Stream Surface Diffusion Network (\cref{fig:pipeline}) that splits processing into two parallel branches: Stream 1 handles global spectral propagation, while Stream 2 performs local high-frequency refinement, as depicted in \cref{fig:EDB}.
\subsection{Preprocessing and Geometric Conditioning}

To relate the input images to the geometry, we render the untextured mesh from three canonical viewpoints, detect facial keypoints on both renders and inputs with RTMPose~\cite{mmpose2020}, and recover extrinsics with EPnP~\cite{lepetit2009ep} and RANSAC~\cite{RANSAC}.
Projecting each vertex into the input views yields image coordinates $P(i)$ and binary visibility masks $V(i)$ that restrict attention to unoccluded regions; per-vertex semantic labels $S \in \{\text{eyes, mouth, face, rest}\}^{|V|}$ come from projected FaRL~\cite{zheng2021farl} segmentation.
 
For the mesh we compute the cotangent Laplacian $L$, mass matrix $M$ and gradient operator $G$, and the $K=300$ smallest eigenpairs of $L$ for the closed-form spectral heat equation, with the fidelity/cost analysis regarding the number of eigenpairs in the supplementary.
Part of the spectrum of eigenpairs is used to compute the Wave Kernel Signature~\cite{aubry2011wave} and normalised positions used to prime the vertex features in the input block.

\subsection{Enhanced Diffusion Block (EDB)}

Prior to intrinsic diffusion processing, noisy vertex attributes are concatenated with bilinearly sampled image features and projected into a latent space. Geometric descriptors derived from WKS and vertex coordinates are injected through affine modulation. Input images are processed by a lightweight multi-scale CNN to provide conditioning features for subsequent cross-attention.

\begin{figure}[tbh]
    \centering
    \includegraphics[width=\columnwidth]{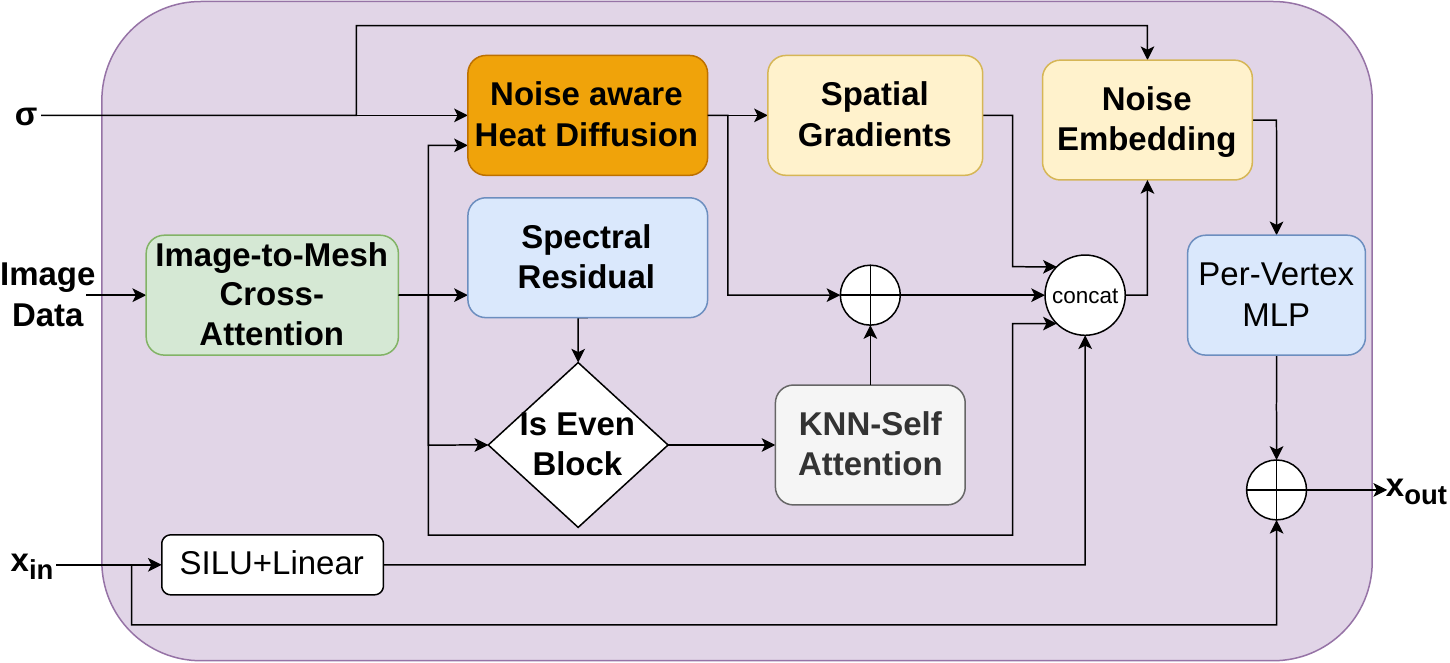}
    \caption{\textbf{Enhanced Diffusion Block (EDB).} Input features $x_{\mathrm{in}}$ are first grounded with multi-view image evidence through image-to-mesh cross-attention. The conditioned features are processed by two complementary parallel streams: noise-aware spectral heat diffusion for global propagation and geodesic KNN self-attention for local refinement. The fused representation is modulated by the diffusion noise level $\sigma$ and projected to $x_{\mathrm{out}}$.}
    \label{fig:EDB}
\end{figure}

The initialised features are then iteratively refined by a stack of EDBs, detailed in \cref{fig:EDB}.
Standard spectral diffusion acts as a low-pass filter, which is counterproductive for synthesising sharp textures. To address this, the EDB replaces monolithic processing with a sequential-to-parallel data flow: it first injects view-dependent appearance cues, and then splits into a dual-stream architecture: one stream for global spectral propagation and a parallel stream for local detail recovery.
Let $H^{(l)}$ denote the vertex features entering layer $l$. The features are first mapped into a hidden space via a SiLU activation and a linear layer, and then updated through the following mechanisms:

\subsubsection{Image-to-Mesh Cross-Attention}
For each visible vertex, $r \times r$ patches are extracted from one set of multi-view CNN feature maps. Those are used in a multi-head cross-attention where the vertex features act as queries, and the sampled patches act as keys and values. This grounds the 3D surface with local 2D image context, producing an appearance-conditioned feature set. A modulation MLP converts the attended features into per-vertex scale and shift updates, while occluded vertices remain unmodified.

\subsubsection{Stream 1: Global Spectral Stream with Anisotropic Gradients}
The global branch propagates information through spectral heat diffusion on the Laplace--Beltrami basis. We extend DiffusionNet by conditioning the per-channel diffusion time on the current noise level $\sigma$, allowing the spatial support of diffusion to adapt throughout denoising. Following DiffusionNet, anisotropic gradient features are incorporated to capture directional surface variation.

\subsubsection{Stream 2: Local High-Frequency Refinement}
Operating in parallel with the global branch, the local stream recovers detail lost through spectral truncation and diffusion smoothing. Refinement is performed using multi-head self-attention over a geodesic $K$-nearest-neighbour graph, preventing information leakage between spatially adjacent but topologically disconnected regions. Region-specific cross-attention, refinement, and output heads are instantiated for the eyes, mouth, face, and remaining head regions.

\subsection{Tangent-Space Output Projection}
Following the sequence of EDBs, a linear projection maps the refined vertex features $H^{(N)}$ back to the albedo or roughness domains.
However, the eye regions demand exceptionally sharp, high-contrast detail that pure vertex-based processing struggles to resolve. To overcome this, we introduce an auxiliary tangent-space refinement branch specifically for the eyes.
For each eye, we compute a local orthonormal tangent basis $(t, b, n)$ based on the region's mean surface normal. Vertex features are projected onto this 2D plane and rasterised onto a $64 \times 64$ grid via inverse distance weighting. A lightweight 2D CNN refines this grid to recover sharp iris and eyelid details. Finally, these features are bilinearly sampled back to the mesh vertices as a residual to the linear output. 

\subsection{Training Objective}
Following the EDM framework, we optimise VISAGE by training the network $D_{\theta}$ to directly reconstruct the clean vertex features via a weighted $L_2$ loss. Noise levels are sampled according to $\ln(\sigma) \sim \mathcal{N}(P_{\text{mean}}, P_{\text{std}}^2)$. To prevent padded input from corrupting batched updates, the loss is computed exclusively over valid mesh vertices via a binary mask $M \in \{0, 1\}^{|\mathcal{V}|}$:
\begin{equation} \label{eq:edm_loss} \mathcal{L} = \mathbb{E}_{\mathbf{x}, \mathbf{n}, \sigma} \left[ \frac{\lambda(\sigma)}{\sum_{v \in \mathcal{V}} M_v} \sum_{v \in \mathcal{V}} M_v \left\| D_\theta(\mathbf{x}_\sigma, \sigma, \mathbf{c})_v - \mathbf{x}_v \right\|_2^2 \right], \end{equation}
where $v$ indexes the vertices, $\mathbf{c}$ denotes the combined geometric and multi-view conditioning signal, and $\lambda(\sigma)$ is the standard EDM weighting function.

\begin{table*}[htb]
    \centering
    \caption{\textbf{Quantitative comparison with state-of-the-art methods on NPHM~\cite{NPHM} and zero-shot generalisation on FaceVerse~\cite{FaceVerse} and i3DMM~\cite{i3DMM}.} VISAGE achieves the best overall perceptual quality across all evaluated datasets, obtaining the best FID and best or tied-best SSIM and LPIPS on each dataset. Best results are \textbf{bold}; second-best are \underline{underlined}. DiffusionNet* denotes our modified baseline with cross-attention conditioning for fair comparison.}
    \label{tab:quantitative_all}
    \adjustbox{width=\textwidth}{%
    \begin{tabular}{lcccccccccccc}
        \toprule
        & \multicolumn{4}{c}{\textbf{NPHM}}
        & \multicolumn{4}{c}{\textbf{FaceVerse}}
        & \multicolumn{4}{c}{\textbf{i3DMM}} \\
        \cmidrule(lr){2-5}
        \cmidrule(lr){6-9}
        \cmidrule(lr){10-13}
        
        \textbf{Model}
        & \textbf{FID} $\downarrow$ & \textbf{SSIM} $\uparrow$ & \textbf{LPIPS} $\downarrow$ & \textbf{PSNR} $\uparrow$
        & \textbf{FID} $\downarrow$ & \textbf{SSIM} $\uparrow$ & \textbf{LPIPS} $\downarrow$ & \textbf{PSNR} $\uparrow$
        & \textbf{FID} $\downarrow$ & \textbf{SSIM} $\uparrow$ & \textbf{LPIPS} $\downarrow$ & \textbf{PSNR} $\uparrow$ \\
        
        \midrule
        
        DiffusionNet*~\cite{DifNet}
        & 22.07 & \underline{0.84} & \underline{0.12} & 20.10
        & 46.72 & 0.80 & 0.19 & 15.74
        & 74.71 & 0.72 & 0.25 & 14.37 \\
        
        Hunyuan3D-2.0-Paint~\cite{Hunyuan3d}
        & \underline{17.66} & 0.81 & \underline{0.12} & \textbf{30.12}
        & \underline{33.30} & 0.79 & \underline{0.17} & \textbf{28.26}
        & 59.41 & 0.80 & \textbf{0.18} & \underline{16.54} \\
        
        MonoNPHM~\cite{MonoNPHM}
        & 19.36 & \underline{0.84} & 0.13 & 23.50
        & 35.61 & \underline{0.81} & 0.19 & 17.34
        & 58.33 & \textbf{0.81} &  0.19 & 15.87 \\
        
        VISAGE
        & \textbf{10.12} & \textbf{0.87} & \textbf{0.08} & \underline{29.40}
        & \textbf{25.79} & \textbf{0.83} & \textbf{0.12} & \underline{26.03}
        & \textbf{55.76} & \textbf{0.81} & \textbf{0.18} & \textbf{16.90} \\
        
        \bottomrule
    \end{tabular}
    }
\end{table*}
\subsection{Training and Data Generation}
\subsubsection{Procedural Training Data Generation}
\label{sec:datagen}

Publicly available 3D head datasets with decomposed PBR material maps are limited in scale and coverage, particularly for full heads including hair and eyes. As noted in \cref{sec:related}, FFHQ-UV-Intrinsics~\cite{Dib_2024_CVPR} supplies specular intensity rather than roughness and covers the face region only, excluding the hair and eye regions our region-specific modules target, while MultiFace~\cite{wuu2022multiface} provides no reflectance decomposition. We therefore construct a synthetic augmentation pipeline and apply it to the NPHM dataset~\cite{NPHM}. Because the original 1.5M-vertex meshes are computationally prohibitive in training, we simplify them to approximately $120$K.

Since NPHM provides only diffuse albedo, we procedurally generate roughness and normal maps using a physically based augmentation pipeline driven by geometric, appearance, and procedural noise cues. This produces diverse skin reflectance and micro-geometry variations spanning a wide range of facial characteristics. As this roughness is generated rather than measured, metrics computed against it assess agreement with a physically motivated prior; \cref{subsubsec:AlbedoDist} evaluates albedo accuracy and albedo--illumination disentanglement independently of it. Full augmentation details and parameter ranges are provided in the supplementary material.

\subsubsection{Network Configuration and Training Setup}
We train the albedo and roughness models on 450 subjects from our synthetically augmented NPHM dataset \cite{NPHM}, evaluating on 37 held-out subjects, with each individual featuring up to 26 distinct expressions. To accommodate their distinct statistical properties, the albedo model (conditioned on RGB images) employs a 9-EDB architecture, while the roughness model (conditioned on HSV space) utilises a lighter 4-EDB architecture. Within the EDM~\cite{EDM} framework, we observe that the model converges better when trained on a higher-noise regime, but introduces residual noise. To resolve this, we employ a sequential cascade training strategy: a base network learning at high noise levels, followed by a refinement network specialised in the lower noise spectra. During inference, these models are chained sequentially.
All models are trained for 120 epochs on four A100 GPUs. Optimiser, learning rate schedule, batch size, architectural capacity (including the 2D CNN modules) and noise distributions are listed in the supplementary material.

\section{Evaluation and Ablation}
We compare VISAGE against state-of-the-art approaches for explicit head texturing using both quantitative and qualitative evaluations and conduct ablation studies to analyse individual design choices. Generated PBR materials are rendered under identical conditions in a physically based rendering engine from four canonical views: three covering the frontal face region and one covering the back of the head. The lighting environment matches the one used for the input images at inference time. Methods that do not predict roughness are rendered using the ground-truth roughness map to isolate differences arising from albedo prediction. We report four complementary image-quality metrics: FID~\cite{NIPS2017_8a1d6947}, SSIM~\cite{wang2004image}, LPIPS~\cite{zhang2018unreasonable}, and PSNR. 
\begin{figure*}[htbp]
\centering
\setlength{\tabcolsep}{1pt}

\begin{tabular}{c c c c c c}
    & \textbf{GT} & \textbf{VISAGE} & \textbf{Hunyuan3D} & \textbf{MonoNPHM} & \textbf{DiffusionNet*} \\
    \multirow{2}{*}{\rotatebox{90}{\textbf{NPHM}}} &
    \includegraphics[width=0.20\textwidth]{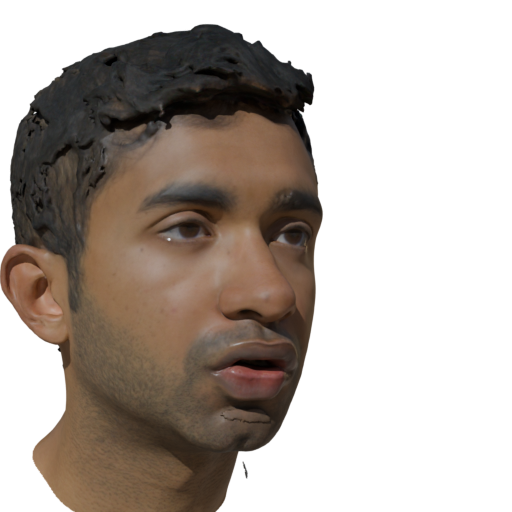} &
    \includegraphics[width=0.20\textwidth]{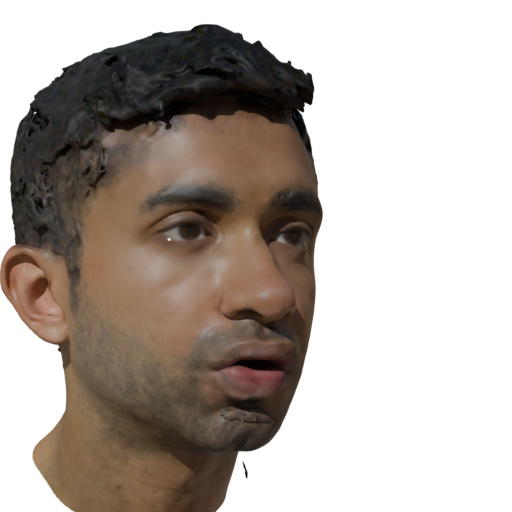} &
    \includegraphics[width=0.20\textwidth]{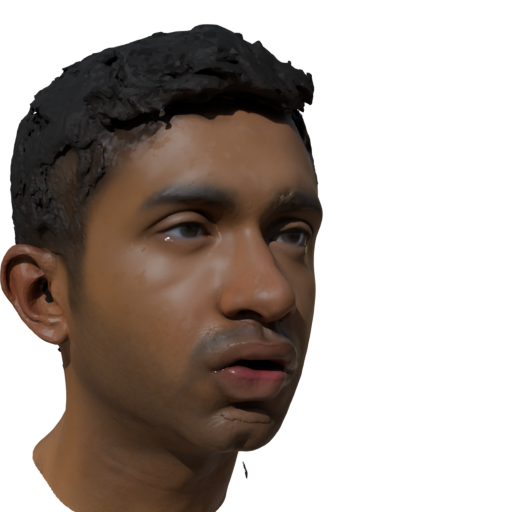} &
    \includegraphics[width=0.20\textwidth]{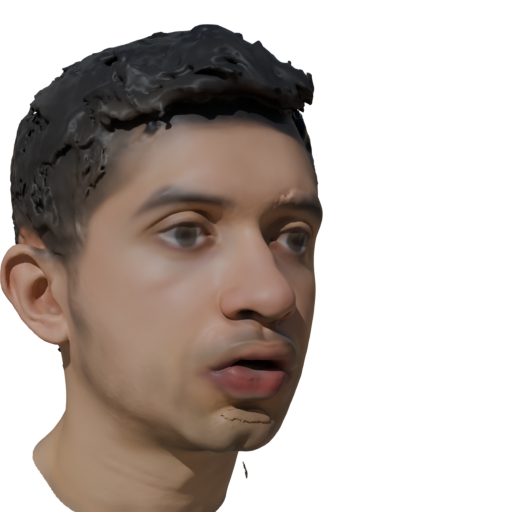} &
    \includegraphics[width=0.20\textwidth]{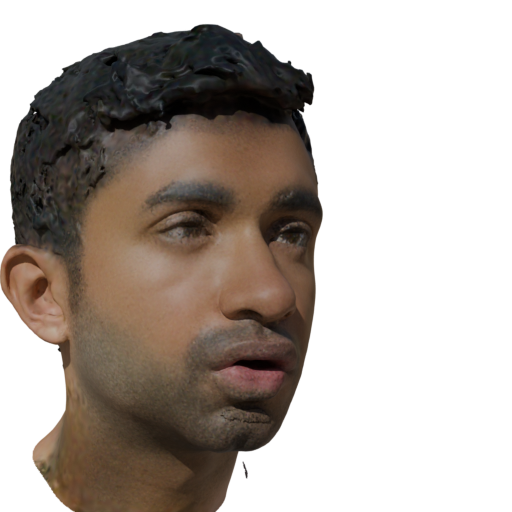} \\
    & %
    \includegraphics[width=0.20\textwidth]{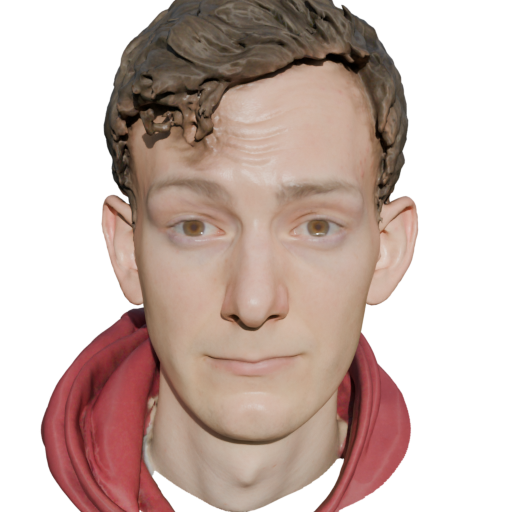} &
    \includegraphics[width=0.20\textwidth]{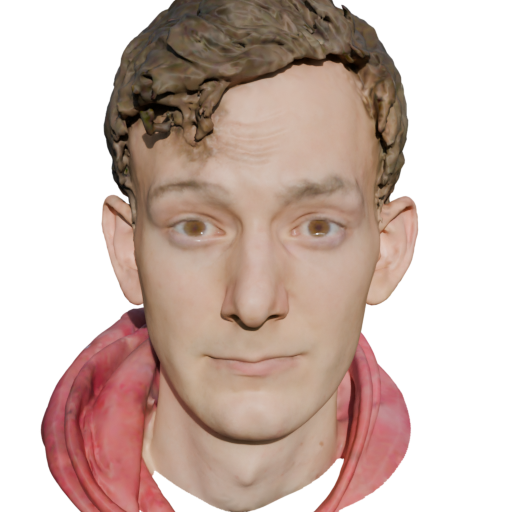} &
    \includegraphics[width=0.20\textwidth]{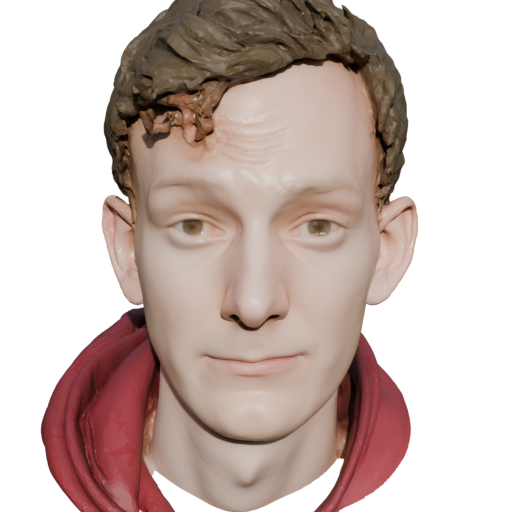} &
    \includegraphics[width=0.20\textwidth]{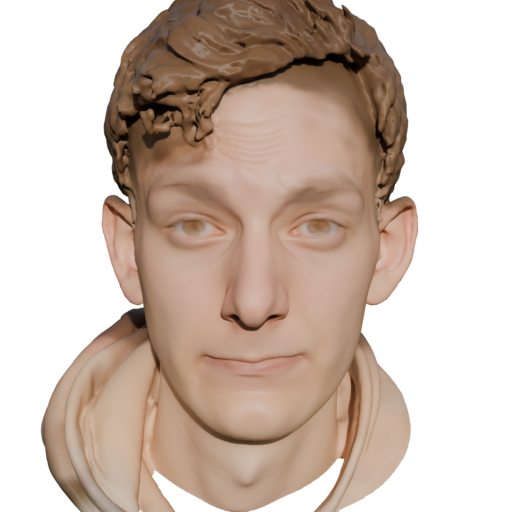} &
    \includegraphics[width=0.20\textwidth]{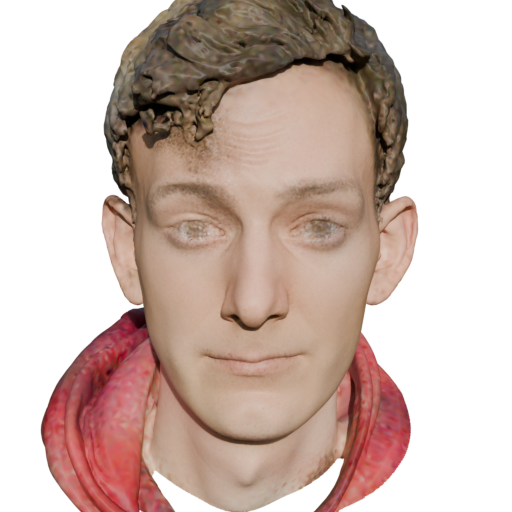} \\

    \multirow{2}{*}{\rotatebox{90}{\textbf{FaceVerse}}} &
    \includegraphics[trim=0 1.5cm 0 0, clip,width=0.20\textwidth]{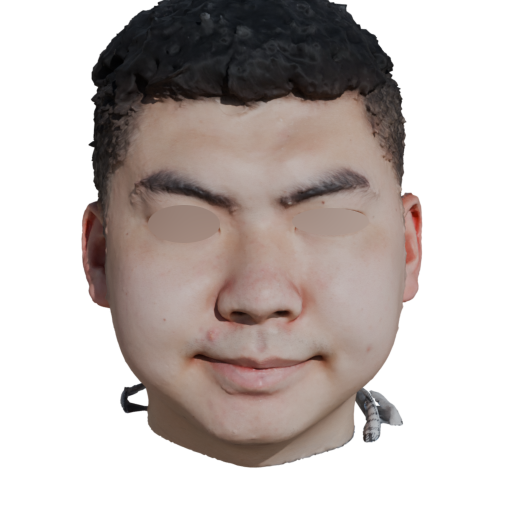} &
    \includegraphics[trim=0 1.5cm 0 0, clip,width=0.20\textwidth]{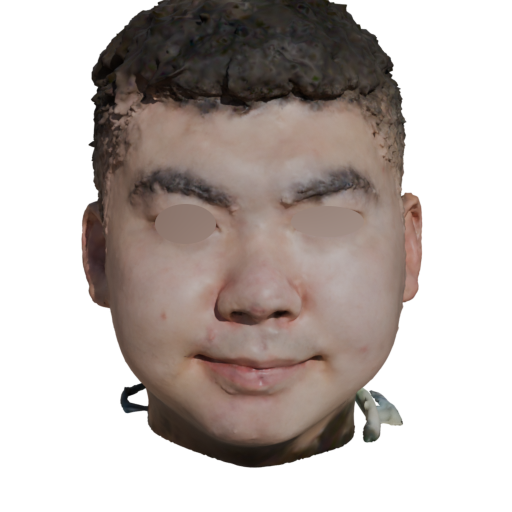} &
    \includegraphics[trim=0 1.5cm 0 0, clip,width=0.20\textwidth]{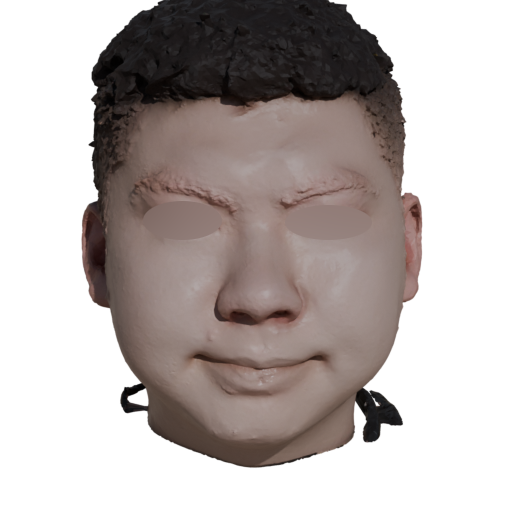} &
    \includegraphics[trim=0 1.5cm 0 0, clip,width=0.20\textwidth]{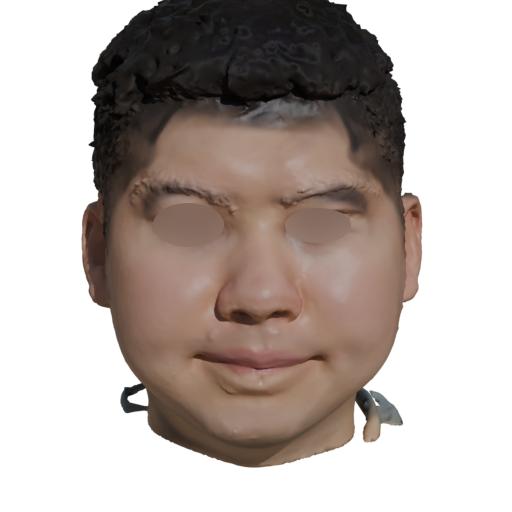} &
    \includegraphics[trim=0 1.5cm 0 0, clip,width=0.20\textwidth]{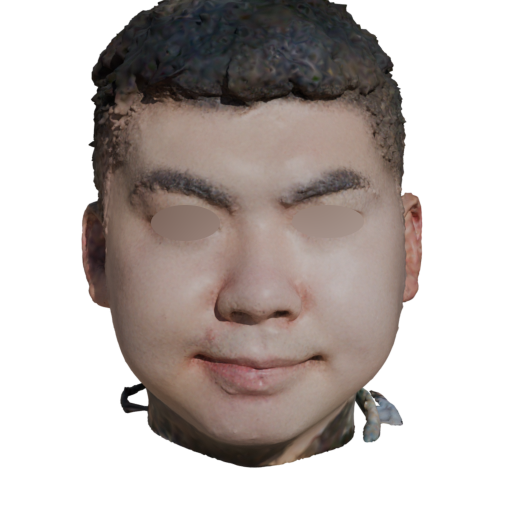} \\
    & %
    \includegraphics[trim=0 0 0 1cm, clip,width=0.13\textwidth]{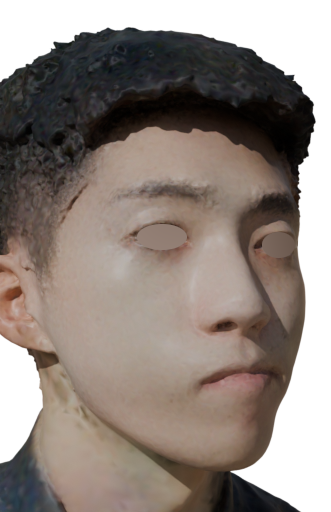} &
    \includegraphics[trim=0 0 0 1cm, clip,width=0.13\textwidth]{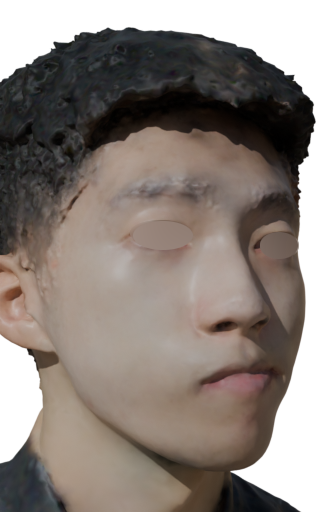} &
    \includegraphics[trim=0 0 0 1cm, clip,width=0.13\textwidth]{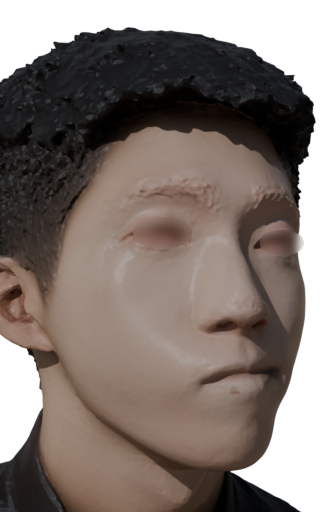} &
    \includegraphics[trim=0 0 0 1cm, clip,width=0.13\textwidth]{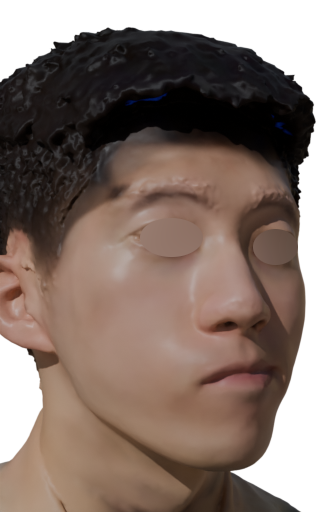} &
    \includegraphics[trim=0 0 0 1cm, clip,width=0.13\textwidth]{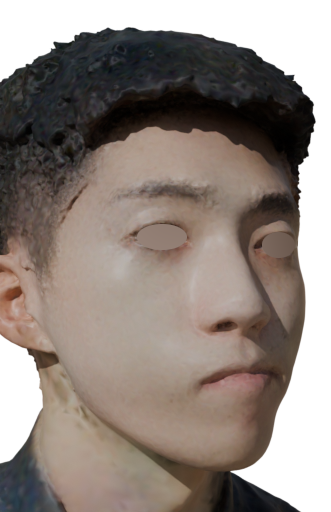} \\

    \multirow{2}{*}{\rotatebox{90}{\textbf{In-the-Wild}}} &
    \includegraphics[width=0.17\textwidth]{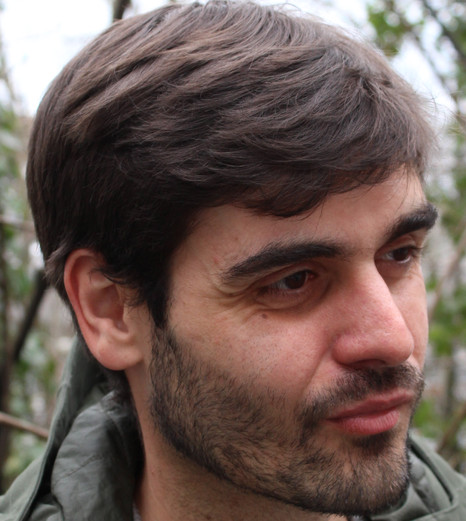} &
    \includegraphics[width=0.17\textwidth]{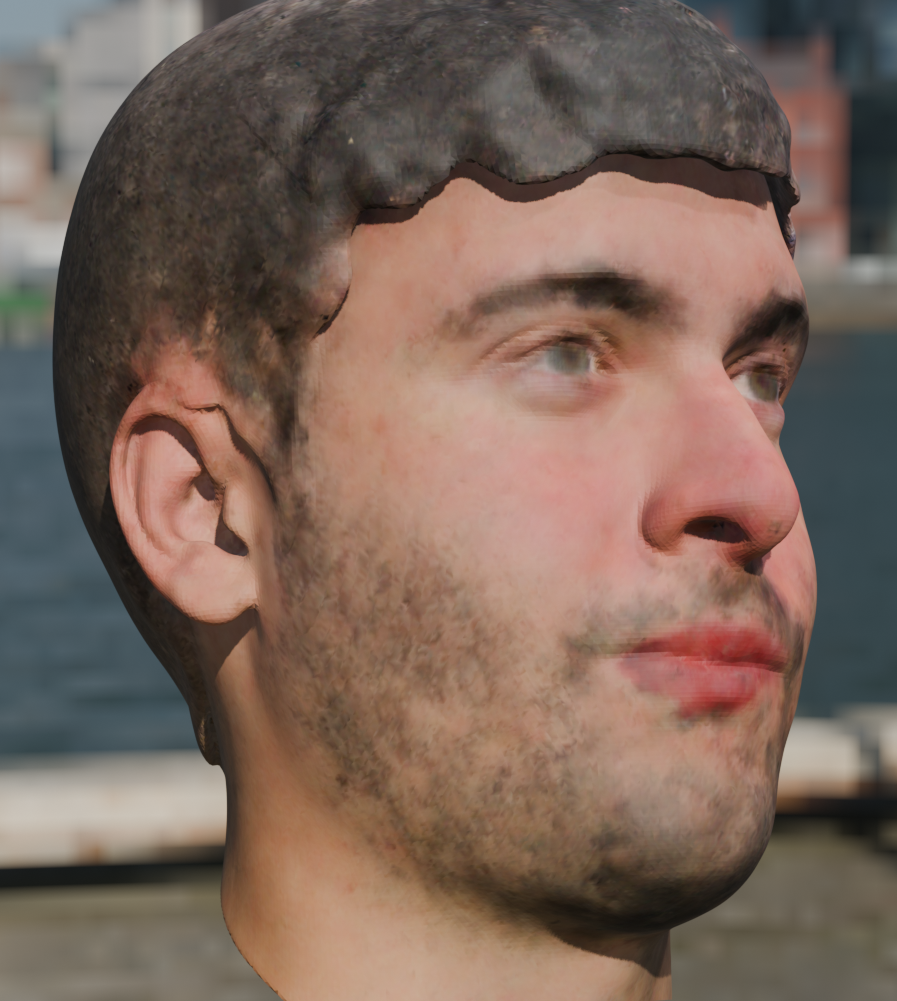} &
    \includegraphics[width=0.17\textwidth]{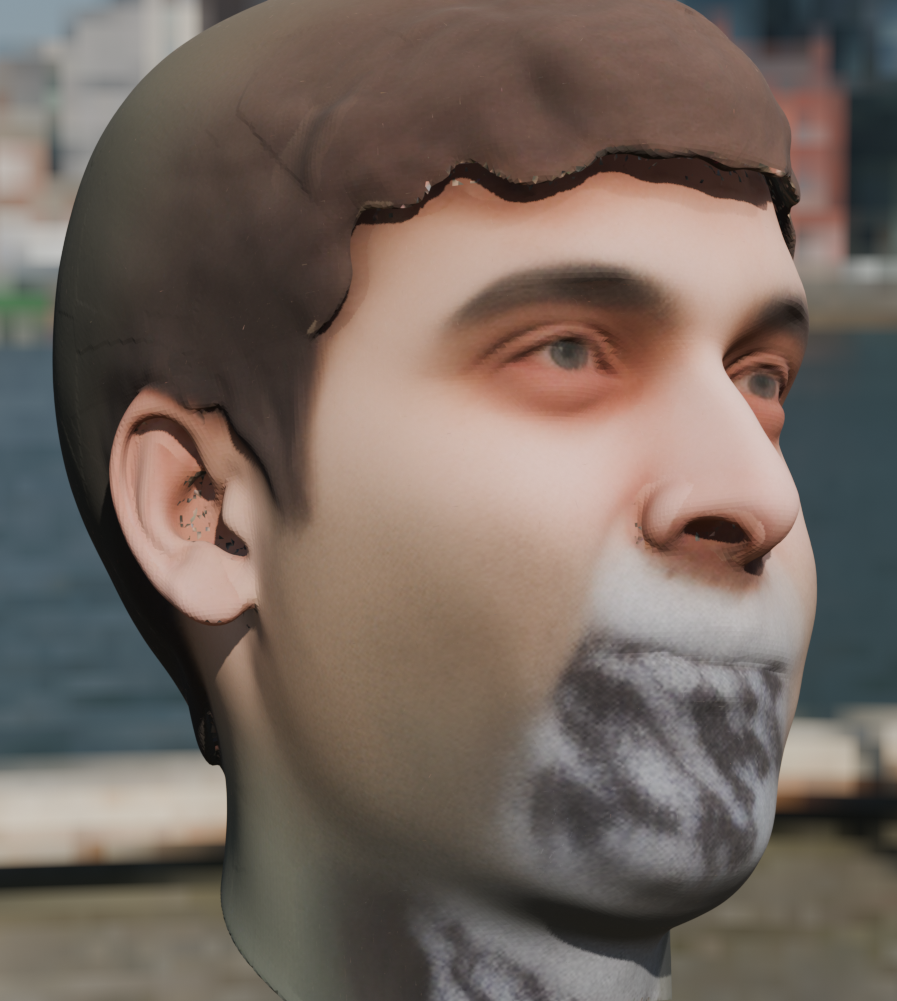} &
    \includegraphics[width=0.17\textwidth]{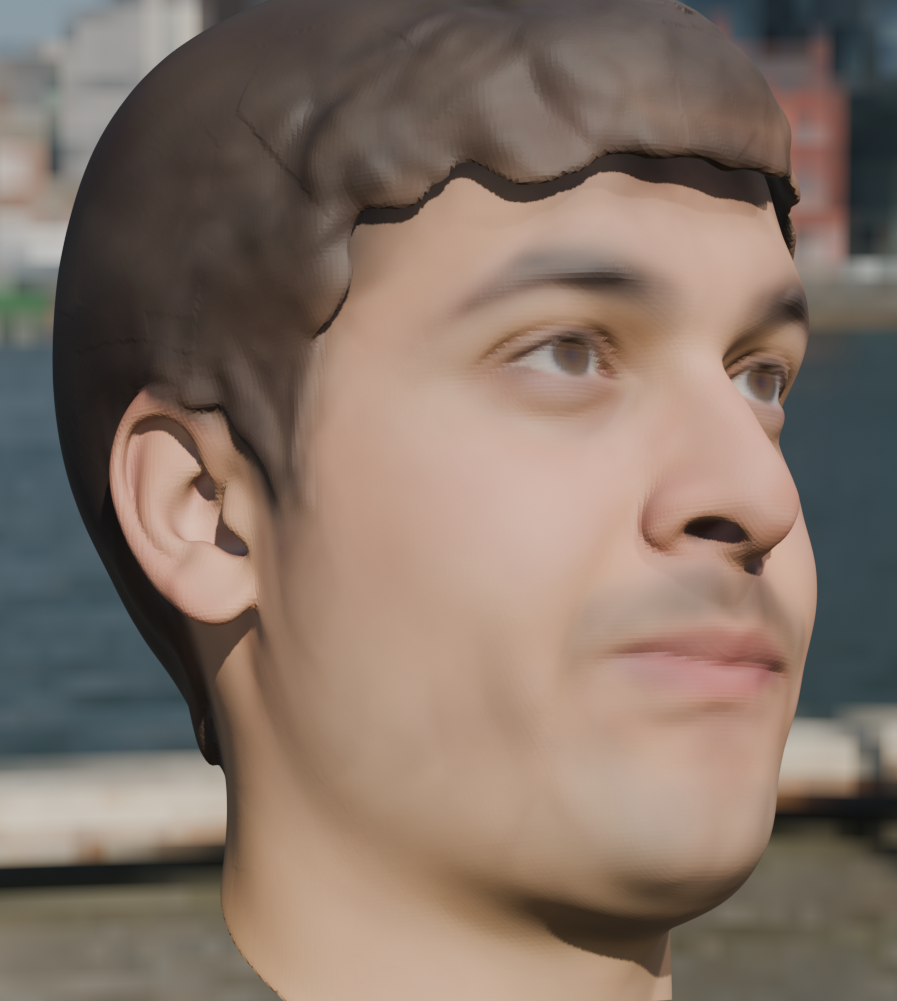} &
    \includegraphics[width=0.17\textwidth]{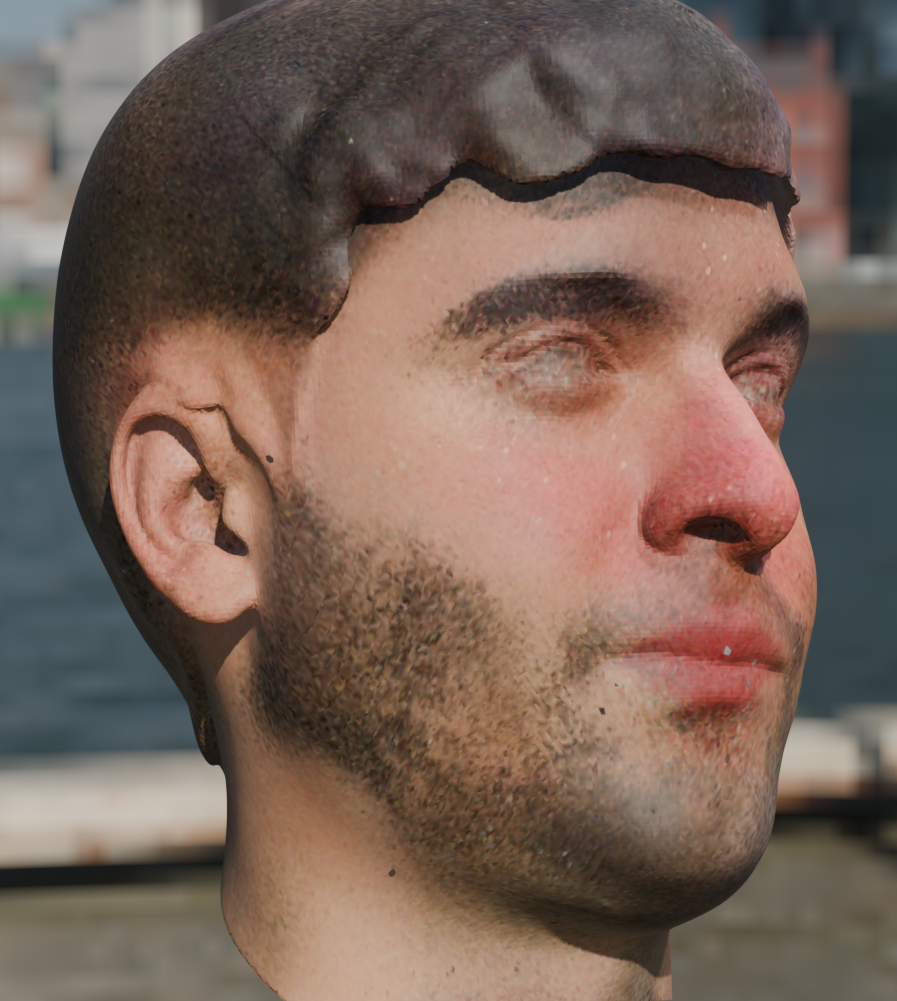} \\
    & %
    \includegraphics[width=0.17\textwidth]{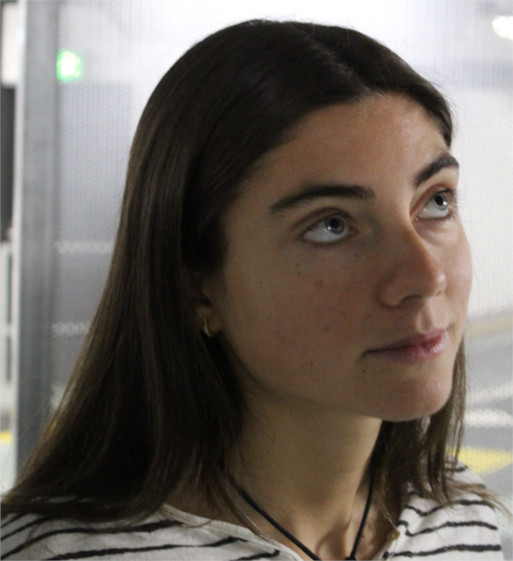} &
    \includegraphics[width=0.17\textwidth]{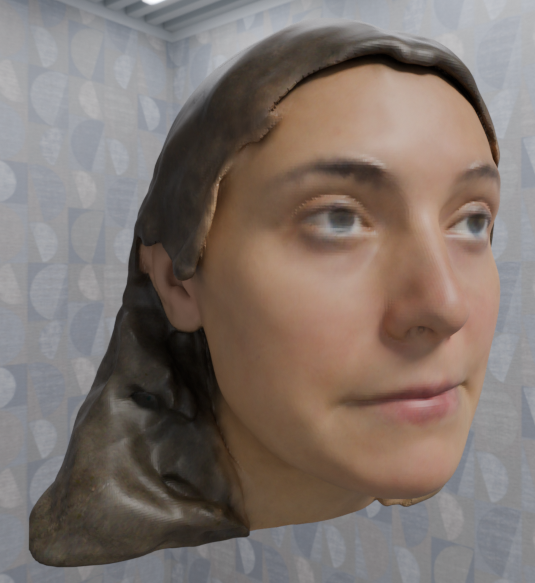} &
    \includegraphics[width=0.17\textwidth]{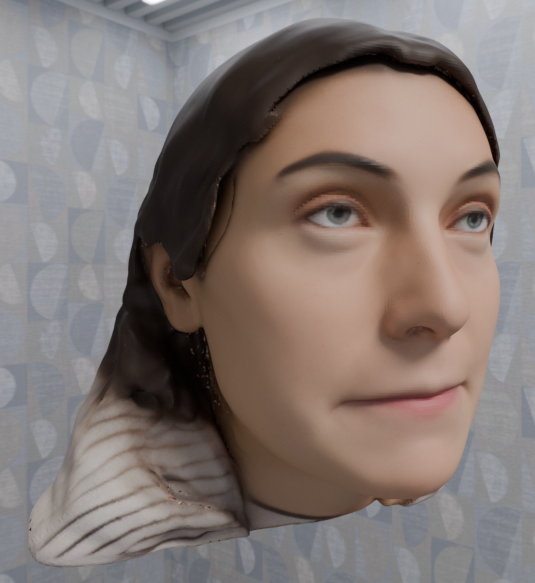} &
    \includegraphics[width=0.17\textwidth]{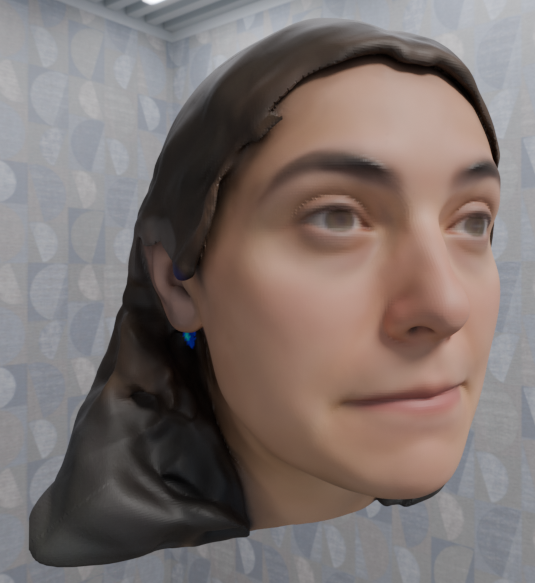} &
    \includegraphics[width=0.17\textwidth]{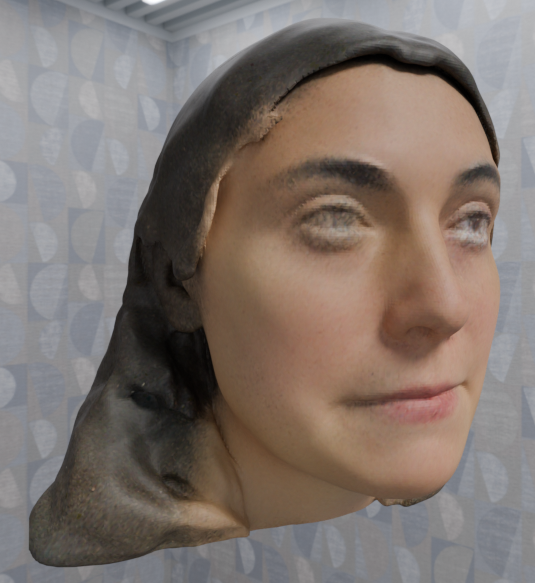} \\
\end{tabular}

\caption{\textbf{Qualitative comparison on NPHM, FaceVerse, and in-the-wild images.} VISAGE better preserves identity-specific appearance, including skin tone, facial hair, and eye colour, while producing more coherent skin and better-resolved eye detail than competing methods. Fine-scale wrinkles around the eyes and mouth (in-the-wild rows) are recovered by none of the compared methods, including ours. Eyes are blurred for FaceVerse~\cite{FaceVerse} examples following the dataset policy.}
\label{fig:NPHMPBR}
\end{figure*}

\subsection{Comparison Against SOTA}
We evaluate VISAGE against three baselines: MonoNPHM~\cite{MonoNPHM}, a neural parametric head model that reconstructs appearance via inverse rendering into neural texture fields; Hunyuan3D-2.0-Paint~\cite{Hunyuan3d}, a state-of-the-art multi-view diffusion pipeline for object texturing, fine-tuned on our augmented NPHM~\cite{NPHM}; DiffusionNet*~\cite{DifNet}, a modified baseline equipped with our cross-attention mechanism to isolate the contribution of our dual-stream architecture. We conduct our primary evaluation on the NPHM dataset and further assess the zero-shot generalisation on the FaceVerse dataset~\cite{FaceVerse} and the test subset of i3DMM~\cite{i3DMM}.
Directly comparable methods are scarce: UV3-TeD~\cite{UV_FREE} surface-diffusion approaches are unconditional, UV-space facial reflectance methods assume a fixed template, and general object-texturing pipelines require domain adaptation; hence our fine-tuning of Hunyuan3D-2.0-Paint. We include MonoNPHM as a reference point rather than a like-for-like generative competitor.
\subsubsection{Quantitative Results}

Our method performs the best across perceptual and structural metrics on NPHM, FaceVerse and i3DMM (\cref{tab:quantitative_all}). On the NPHM dataset, VISAGE improves the FID score by 42.7\% and reduces LPIPS error by 33.3\% compared to the next-best baseline. Similar performance gains are observed in zero-shot results on FaceVerse, where we outperform the next-best model by 22.5\% in FID. This zero-shot generalisation extends to the i3DMM, where VISAGE achieves the lowest FID (55.76) and tied-best SSIM (0.81), outperforming both DiffusionNet* and Hunyuan3D-2.0-Paint while matching the best LPIPS. While Hunyuan3D-2.0-Paint retains an advantage in PSNR except on i3DMM, our approach remains highly competitive, while significantly improving perceptual fidelity.

\subsubsection{Qualitative Results}
\cref{fig:NPHMPBR} presents a qualitative comparison on the NPHM dataset, zero-shot visualisations on the FaceVerse dataset and in the wild. VISAGE reconstructs facial hair and subject-specific eye colour, which most competing approaches over-smooth or hallucinate. Our approach also achieves better skin-colour fidelity; MonoNPHM washes out darker complexions, while Hunyuan3D-2.0-Paint exaggerates skin tones in both directions. Furthermore, while the base DiffusionNet* captures coarse appearance, it produces a grainy texture and struggles with detailed regions like the eyes. In contrast, our dual-stream architecture produces more coherent skin appearance and eye detail, recovering mid-frequency variation such as localised blemishes and beard detail, while pore-level structure and fine expression wrinkles remain beyond the reach of all compared methods at this input sparsity.
On in-the-wild Cafca~\cite{buehler2024cafca} images with MonoNPHM-estimated geometry, VISAGE produces richer skin detail and fewer artefacts than the baselines, particularly at the eyes and hairline.

\subsubsection{Perceptual User Study}
\begin{figure}
    \centering
    \includegraphics[trim=0 0 0 0.5cm, clip,width=\columnwidth]{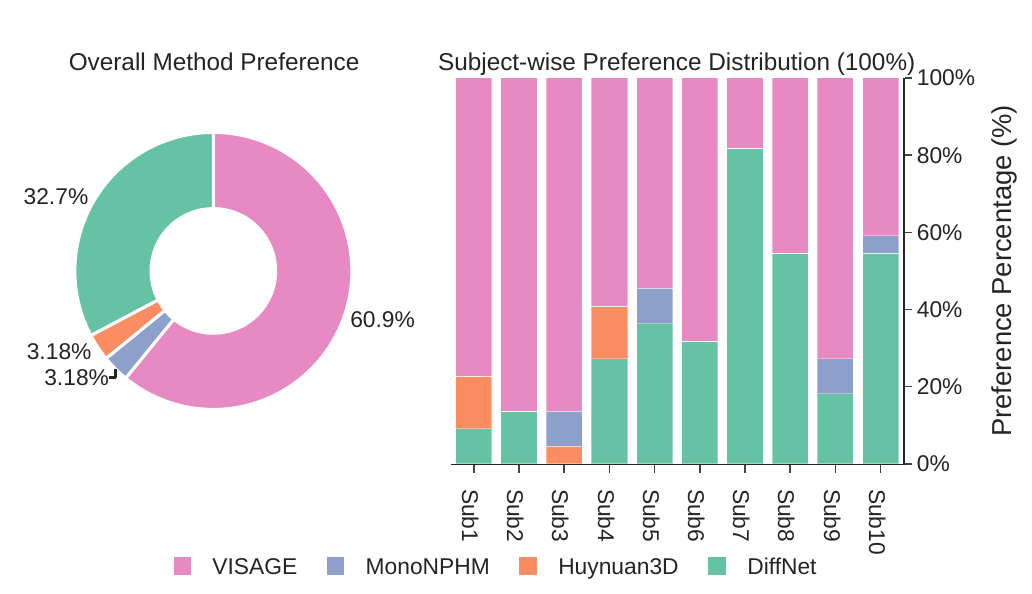}
    \caption{\textbf{Perceptual user study.} Left: overall preference across all participants. Right: subject-wise preference distribution. VISAGE is preferred over competing methods in the majority of trials.}
    \label{fig:user_study}
\end{figure}
To complement our quantitative metrics, we conduct a blind user study with 22 participants comparing VISAGE to baseline methods. We randomly sample 5 identities from NPHM and 5 identities from the zero-shot FaceVerse dataset. We render each method under the same lighting and viewpoint. Each participant is shown results from all methods in randomised order across 10 test subjects and selects the one closest to ground truth. The study is fully anonymous and counterbalanced to avoid ordering bias. As shown in \cref{fig:user_study}, our method is consistently preferred by human raters across most subjects and achieves the highest overall preference score, indicating a consistent preference over the compared baselines.

\subsubsection{Efficiency} 
VISAGE uses only 7.22M parameters in its core generative network and requires 5.6 GB peak GPU memory, while synthesising textures in 104.9\,s. Detailed comparisons with competing methods and preprocessing costs are provided in the supplementary material.
\subsection{Ablation Studies}
To comprehensively evaluate the VISAGE framework, our ablation study includes an analysis of the network component impact, its sensitivity to the number of input views, mesh vertex resolution and illumination disentanglement. Due to computational constraints, all the models are trained on only the lower noise regime. The supplementary material adds geometry robustness and a predicted-vs-constant roughness comparison.
\subsubsection{Impact of Network Components}
\begin{figure}[t]
    \centering
    \begin{subfigure}[b]{0.49\columnwidth}
        \centering
        \includegraphics[width=1\linewidth]{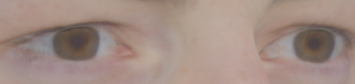}
        \caption{Ground truth}
        \label{fig:GT1}
    \end{subfigure}
        \begin{subfigure}[b]{0.49\columnwidth}
        \centering
        \includegraphics[width=1\linewidth]{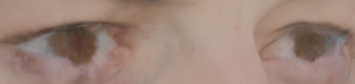}
        \caption{No semantic region splitting}
        \label{fig:NoRegion}
    \end{subfigure}
    \\
    \begin{subfigure}[b]{0.49\columnwidth}
        \centering
        \includegraphics[width=1\linewidth]{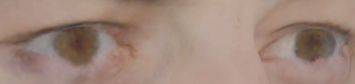}
        \caption{No tangent refinement branch}
        \label{fig:NoTanget2}
    \end{subfigure}
        \begin{subfigure}[b]{0.49\columnwidth}
        \centering
        \includegraphics[width=1\linewidth]{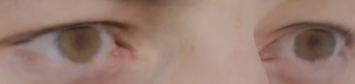}
        \caption{VISAGE}
        \label{fig:Comb2}
    \end{subfigure}
    \caption{\textbf{Effect of region specialisation and tangent refinement.} Removing semantic region splitting degrades iris structure, while removing tangent-space refinement produces blurry eye details. Combining both components restores sharp, realistic eye appearance.}
    \label{fig:Ablatation}
\end{figure}

Disabling \textit{Semantic Region Splitting} forces a shared global MLP across all vertices, failing to capture the unique statistical distributions of the eyes and teeth and severely degrading realism (\cref{fig:Ablatation}); omitting the \textit{Tangent-Space Output Projection} bypasses 2D CNN refinement, yielding blurry irises. Both are quantitatively minor given the small surface area of the eyes (\cref{tab:ablation}) but perceptually noticeable.

Removing the \textit{K-Nearest-Neighbour} stream forces reliance on the EDB's MLP alone, causing grainy artefacts and loss of sharpness (see supplementary material), and removing \textit{Noise-Level Conditioning} prevents the network from adapting its spatial reach, causing texture bleeding at occluded regions and boundaries (\cref{tab:ablation}).

\begin{table}[t]
    \centering
    \caption{\textbf{Ablation on key architectural components.} Each proposed component contributes to reconstruction quality, removing any key module consistently degrades the quality, especially noticeable in the PSNR and FID scores.}
    \label{tab:ablation}
    \adjustbox{width=\columnwidth}{%
    \begin{tabular}{lcccc}
        \toprule
        \textbf{Model} & \textbf{FID} $\downarrow$ & \textbf{SSIM} $\uparrow$ & \textbf{LPIPS} $\downarrow$ & \textbf{PSNR} $\uparrow$ \\
        \midrule
        VISAGE & \textbf{27.74} & \textbf{0.85} & \textbf{0.10} & \textbf{24.72} \\
        w/o High-frequency stream & 36.40 & 0.84 & 0.13 & 21.70 \\
        w/o Noise level in heat diffusion & 35.60 & 0.82 & 0.12 & 20.91 \\
        w/o Region splitting & 31.62 & 0.84 & 0.11 & 23.60 \\
        w/o Tangent refinement & 29.26 & 0.84 & 0.11 & 24.53 \\
        \bottomrule
    \end{tabular}
    }
\end{table}

\begin{figure}[htbp]
    \centering
    \begin{subfigure}[b]{0.49\columnwidth}
        \centering
        \includegraphics[width=\columnwidth]{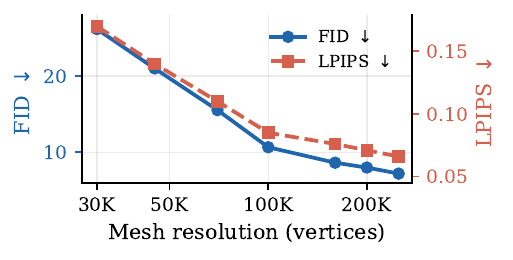}
        \caption{}
        \label{fig:res_per}
    \end{subfigure}
    \hfill
    \begin{subfigure}[b]{0.49\columnwidth}
        \centering
        \includegraphics[width=\columnwidth]{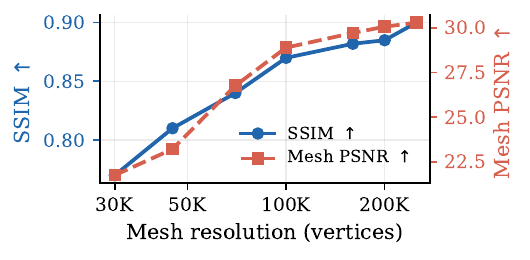}
        \caption{}
        \label{fig:res_struc}
    \end{subfigure}

    \begin{subfigure}[b]{\columnwidth}
        \centering
        \includegraphics[width=\columnwidth]{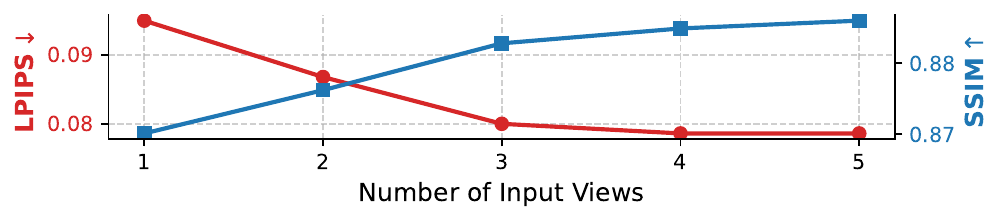}
        \caption{}
        \label{fig:InputImages}
    \end{subfigure}

    \caption{\textbf{Scaling behaviour.} (a,b) Appearance quality against
    mesh resolution: the model trained at 120K vertices, evaluated without
    retraining on NPHM~\cite{NPHM} decimated to 30K--250K vertices. The
    intrinsic operators depend on the surface rather than a fixed
    tessellation, so a single set of weights covers the whole range and
    resolution can be traded against computational cost at inference time.
    (c) Effect of the number of conditioning images, with diminishing
    returns beyond three views.}
    \label{fig:scaling}
\end{figure}
\begin{figure}[t]
    \centering

    \resizebox{\columnwidth}{!}{%
    \begin{tabular}{lccc}
        \toprule
        \textbf{Region} &
        \textbf{Spread $\Delta E\downarrow$} &
        \textbf{GT $\Delta E\downarrow$} &
        \textbf{Observed $\Delta E$} \\
        \midrule
        All   & $1.56 \pm 0.22$ & $4.41 \pm 0.44$ & $9.38$ \\
        Face  & $1.53 \pm 0.18$ & $3.35 \pm 0.26$ & $11.06$ \\
        Hair  & $1.65 \pm 0.39$ & $6.62 \pm 1.09$ & $5.94$ \\
        Eyes  & $1.35 \pm 0.11$ & $3.68 \pm 0.29$ & $10.94$ \\
        Mouth & $1.43 \pm 0.15$ & $3.16 \pm 0.55$ & $11.34$ \\
        \bottomrule
    \end{tabular}%
    }

    \includegraphics[width=\columnwidth]{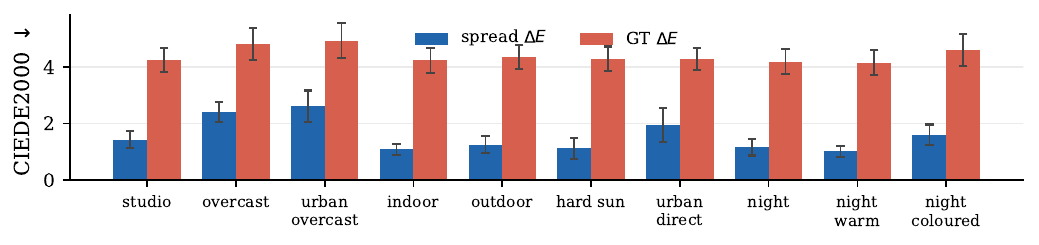}

    \caption{\textbf{Albedo illumination invariance.}
    Results on the NPHM~\cite{NPHM} validation split under 10 environment maps.
    \emph{Spread} is the mean CIEDE2000~\cite{Luo2001CIEDE2000} distance between each illumination's predicted albedo and the per-vertex mean prediction; \emph{GT} is the distance to ground-truth albedo, and \emph{Observed} is the corresponding variation in the input observations.
    Table: regional mean $\pm$ std across validation subjects. Plot: results by environment map.
    Predicted variation remains substantially below observed variation.}
    \label{fig:illumination_invariance}
\end{figure}

\subsubsection{Varying Number of Input Images and Mesh Vertex Resolution}
\label{subsubsec:Resolution}
We next analyse the impact of the number of conditioning views. We vary the number of input images from 1 to 5 while keeping geometry and rendering settings fixed. Remarkably, VISAGE already achieves competitive performance with a single input image: most regions are reconstructed faithfully, with errors concentrated in occluded areas such as the back of the head and hairline. As the number of views increases, the network gains additional spatial context, leading to improvements in LPIPS and SSIM. 
However, \cref{fig:scaling} shows clear diminishing returns beyond three images. We therefore adopt three views as a practical default, while single-image inference remains an effective option for minimal-input scenarios.
We further examine how appearance quality depends on mesh resolution, which is the natural counterpart to texture resolution in a UV-based pipeline. Keeping the model trained at $120$K vertices fixed, we evaluate it on the same subjects decimated to between $30$K and $250$K vertices (\cref{fig:scaling}). Quality is clearly resolution-dependent, as expected for a vertex-bound representation: FID improves from $26.2$ at $30$K to $7.2$ at $250$K. Worth noting is that the model transfers across the entire range without retraining, despite having seen only a single resolution during training; the intrinsic operators depend on the surface rather than on a fixed tessellation, so no resolution-specific model is required. Mesh resolution therefore acts as a quality-versus-cost control at inference time: a practitioner may decimate for speed or refine for fidelity using the same weights, rather than being restricted to the resolution the model was trained at.

\subsubsection{Illumination Disentanglement}
\label{subsubsec:AlbedoDist}

We render NPHM~\cite{NPHM} validation subjects under 10 held-out studio, outdoor, indoor, and night environments, and measure CIEDE2000~\cite{Luo2001CIEDE2000} variation across predicted albedos and error to ground-truth albedo. As shown in \cref{fig:illumination_invariance}, predicted albedo variation is low across illuminations ($\Delta E=1.56$ overall), removing $83\%$ of the observed variation ($\Delta E=9.38$); on the face, it removes $86\%$ ($1.53$ vs.\ $11.06$). Spread remains low across environments, while ground-truth error is stable, suggesting that remaining error is largely independent of illumination. Errors are higher for hair than for the face ($\text{GT }\Delta E=6.62$ vs.\ $3.35$).

\section{Limitations}
Several limitations remain. First, albedo and illumination remain partially entangled under difficult lighting: strong cast shadows and coloured light are occasionally absorbed into the predicted albedo, most noticeably in hair; \cref{subsubsec:AlbedoDist} quantifies this via the variance of predicted albedo across illuminations of the same subject.
Second, appearance resolution is bound to vertex density rather than decoupled from it as in UV-mapped assets; quality improves with mesh resolution (\cref{subsubsec:Resolution}), but pore-level detail as carried by 4K textures remains out of reach.
Third, roughness supervision is procedurally generated, so we validate its consistency and rendering effect but not its physical accuracy, as no public full-head reflectance dataset exists.
Fourth, since image supervision covers only frontal and lateral views, heavily occluded regions such as the back of the head rely largely on the diffusion prior and can show subtle luminance discontinuities.
Finally, although our augmentation pipeline generates normal maps, we train only for albedo and roughness; jointly predicting high-frequency normals is a natural extension.

\section{Conclusion}
We presented VISAGE, a surface-based diffusion framework that synthesises PBR-compatible per-vertex albedo and roughness on arbitrary head meshes from sparse RGB images, without requiring a consistent UV parameterisation. Combining global spectral propagation with local geodesic KNN refinement under visibility-aware multi-view conditioning, VISAGE improves FID by \textbf{42.7\%} over the next-best baseline on NPHM~\cite{NPHM}, with zero-shot gains on FaceVerse~\cite{FaceVerse} (\textbf{22.5\%}) and i3DMM~\cite{i3DMM} (\textbf{4.4\%}), using a 7.2M-parameter core network, and is preferred in a blind perceptual study. Ablations identify the KNN refinement stream and noise-level-aware diffusion as most critical for boundary fidelity and occluded regions. Despite training only on procedurally augmented synthetic data, the model transfers to real-world images.

\section*{Acknowledgements}
This work has been partially funded by the EU project IRIS-XR (GA: Nr 101298672) and by the EU project VICTOR-XR (GA: Nr 101298864)

{
    \small
    \bibliographystyle{ieeenat_fullname}
    \bibliography{main}
}
\clearpage
\appendix
\maketitlesupplementary
\section{Overview}
\label{sec:overview}

This document is the technical appendix for \textit{VISAGE: Vertex-based Intrinsic Surface Appearance GEneration via Dual-Stream Diffusion for Arbitrary 3D Heads}. It is organised as follows.

\paragraph{\Cref{supp:augmentation} -- Augmentation Pipeline.}
We provide the full technical description of our physics-based PBR augmentation
pipeline, including per-region synthesis
strategies, algorithms and parameter ranges. 
This supplements Sec.~3.5.1. of the main paper, which defers its complete
specification to this appendix.

\paragraph{\Cref{supp:ArchitecturalDetails} -- Architectural Details.}
We detail the components of VISAGE, abbreviated in the main paper due to space constraints. This includes the full CNN architectures for image pre-processing and tangent-space refinement (Secs.~3.1., 3.2. and 3.3. of the main paper), a table of all training hyperparameters, the noise scheduling, and the inference sampling strategy underlying our two-stage EDM training setup (Secs.~3.4 and 3.5.2. of the main paper), and an efficiency analysis of the complete pipeline.

\paragraph{\Cref{supp:AddAblations} -- Additional Ablations and Visualizations.}
We quantify robustness to noisy input geometry, isolate the contribution of the
predicted roughness against a constant-roughness fallback, and provide the
remaining qualitative ablation comparisons deferred from Sec.~4.2.1 of the main
paper, isolating the contributions of the high-frequency stream, noise-level
conditioning, and semantic region splitting. We also present representative
failure cases that characterise the current limitations of VISAGE.

\section{Augmentation Pipeline} \label{supp:augmentation}
\begin{figure}[htb]
    \centering
    \begin{subfigure}[b]{0.25\linewidth}
        \includegraphics[width=\linewidth]{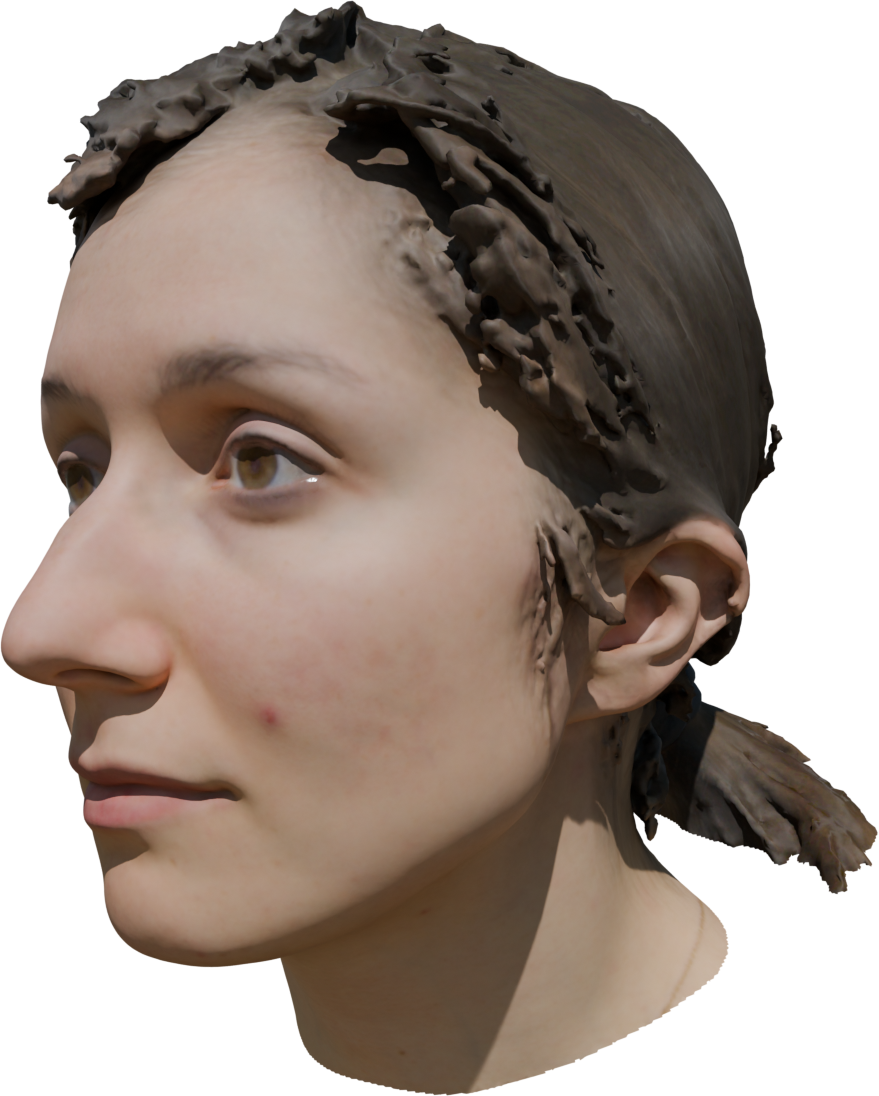}
    \end{subfigure}\hfill
    \begin{subfigure}[b]{0.25\linewidth}
        \includegraphics[width=\linewidth]{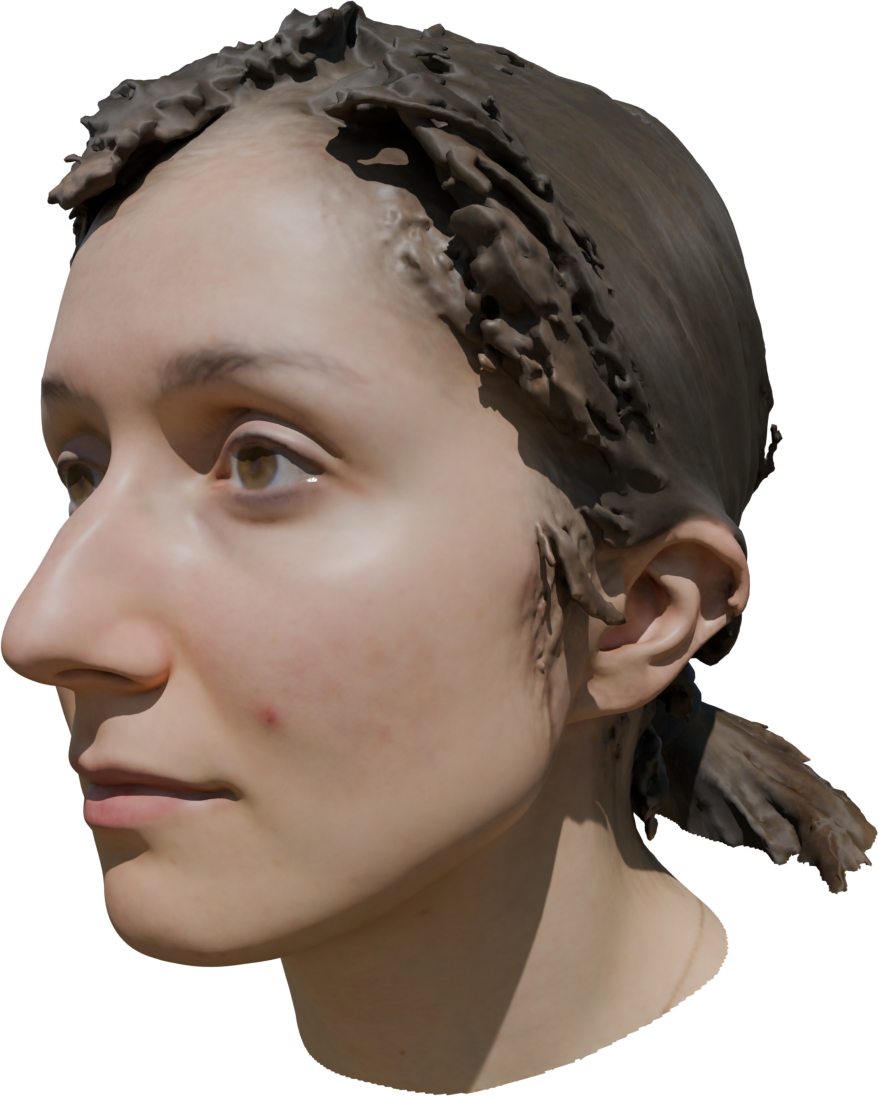}
    \end{subfigure}\hfill
    \begin{subfigure}[b]{0.25\linewidth}
        \includegraphics[width=\linewidth]{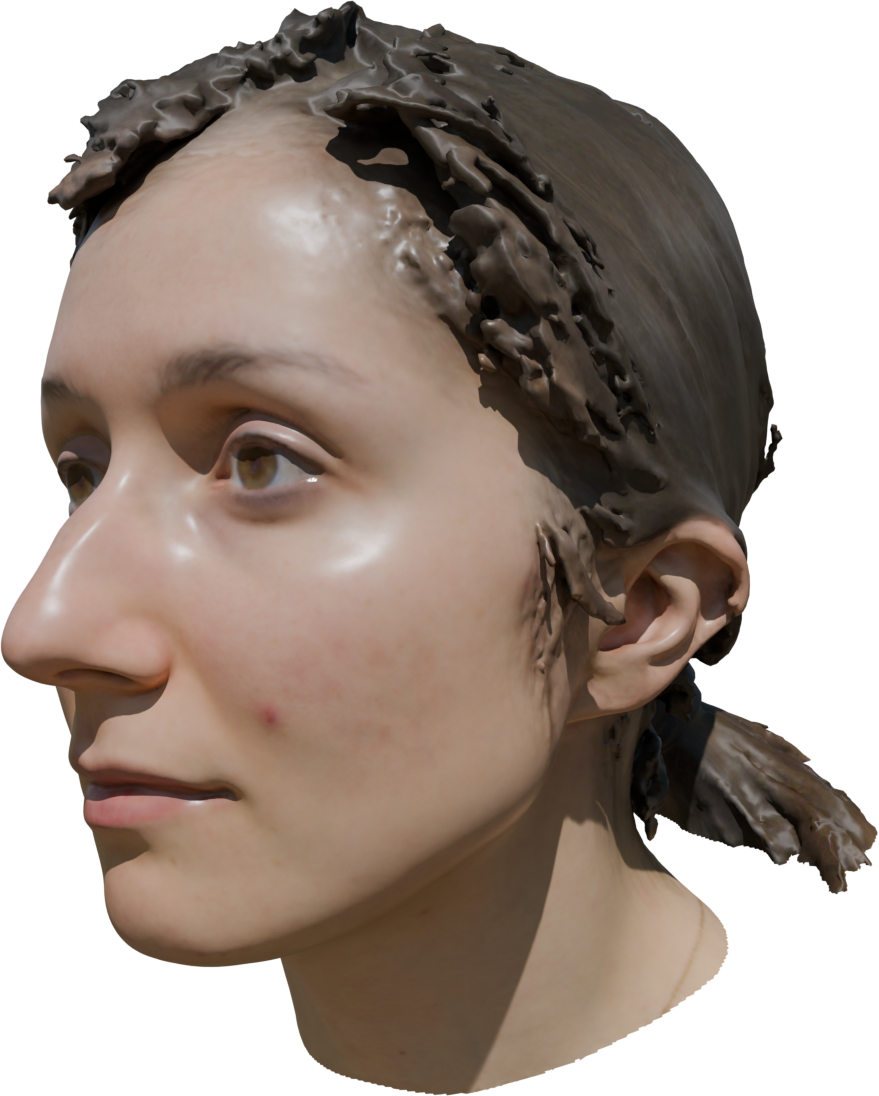}
    \end{subfigure}
    
    \vspace{0.1cm} %
    
    \begin{subfigure}[b]{0.32\linewidth}
        \includegraphics[width=\linewidth]{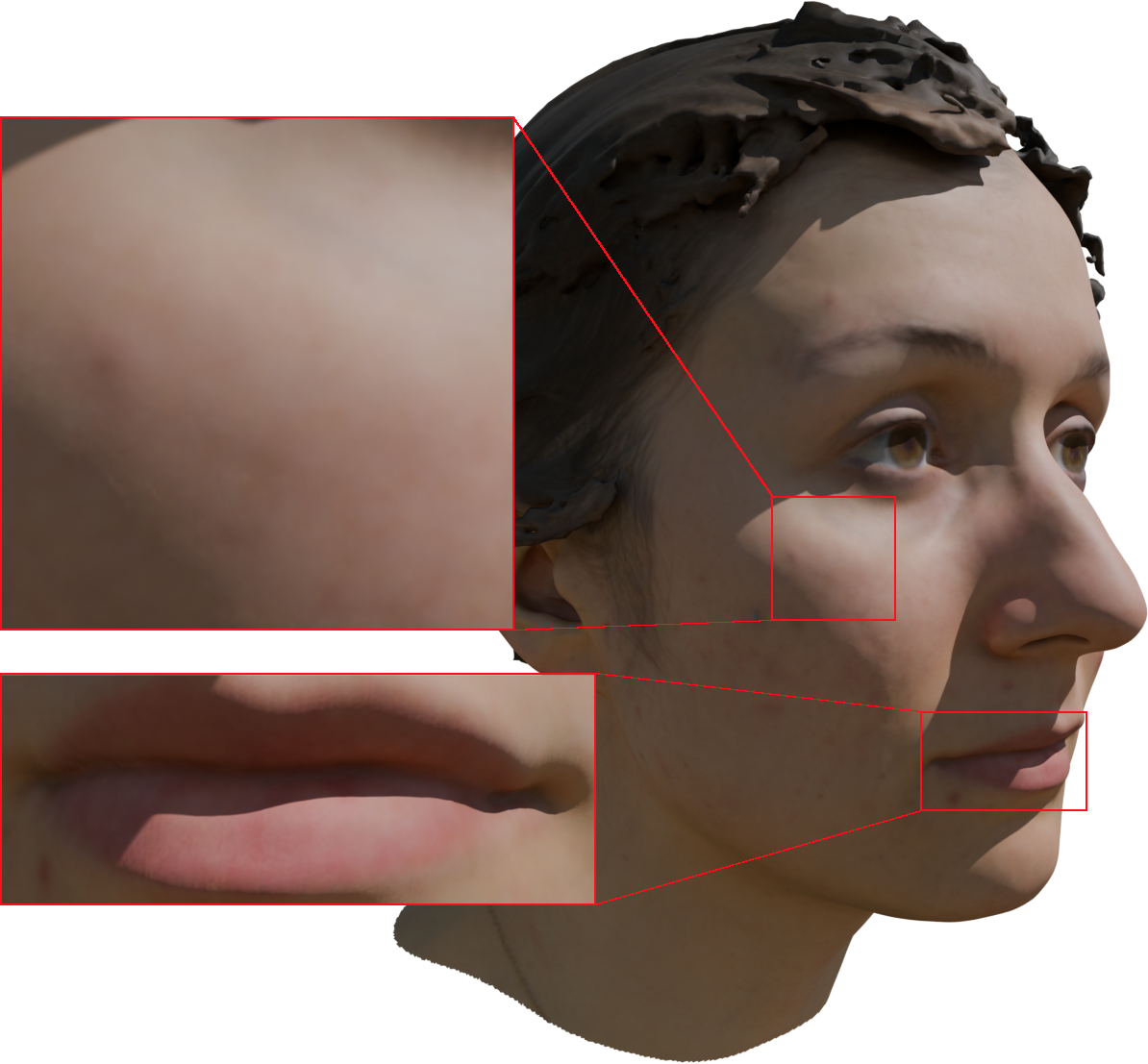}
    \end{subfigure}\hfill
    \begin{subfigure}[b]{0.32\linewidth}
        \includegraphics[width=\linewidth]{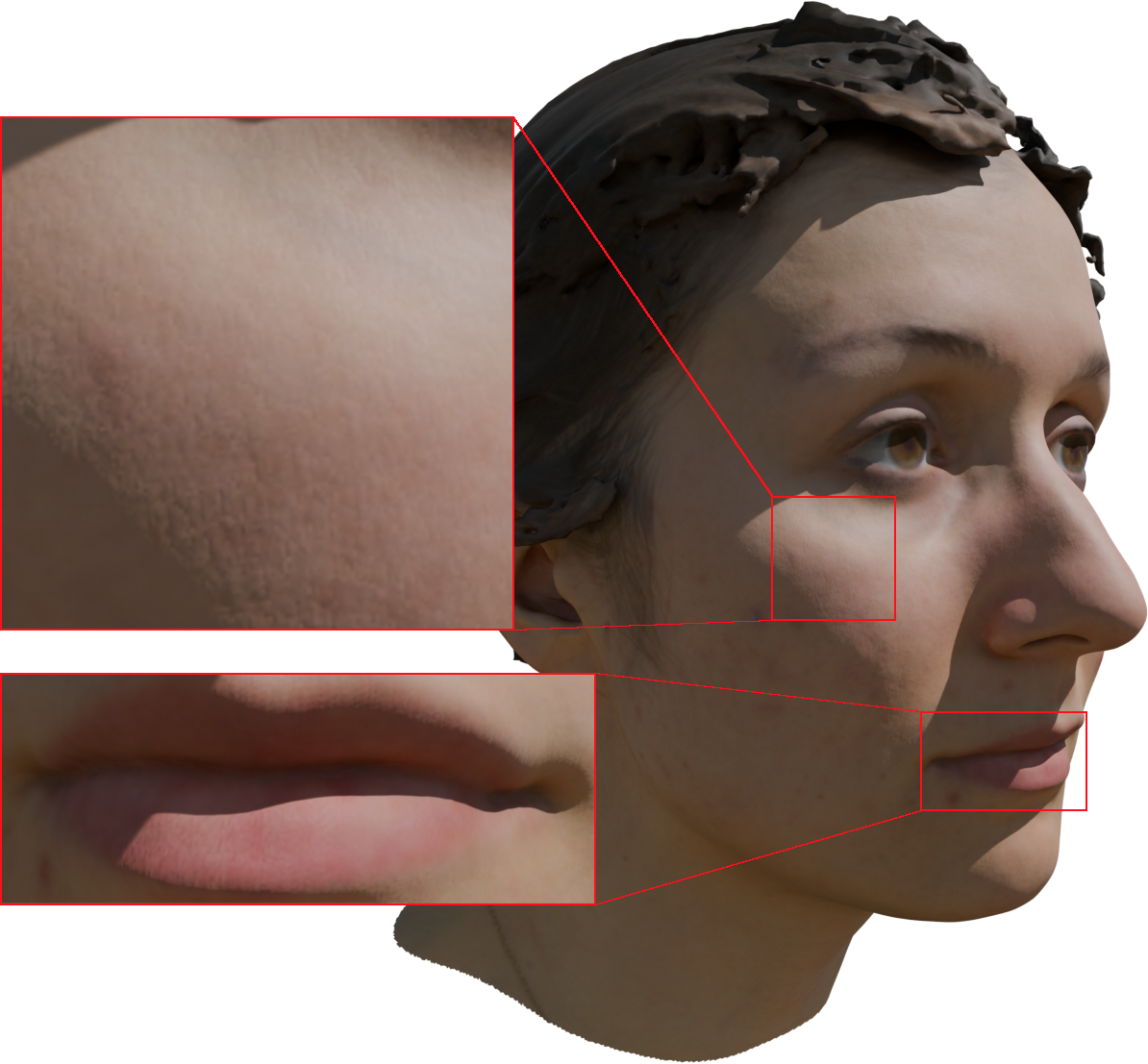}
    \end{subfigure}\hfill
    \begin{subfigure}[b]{0.32\linewidth}
        \includegraphics[width=\linewidth]{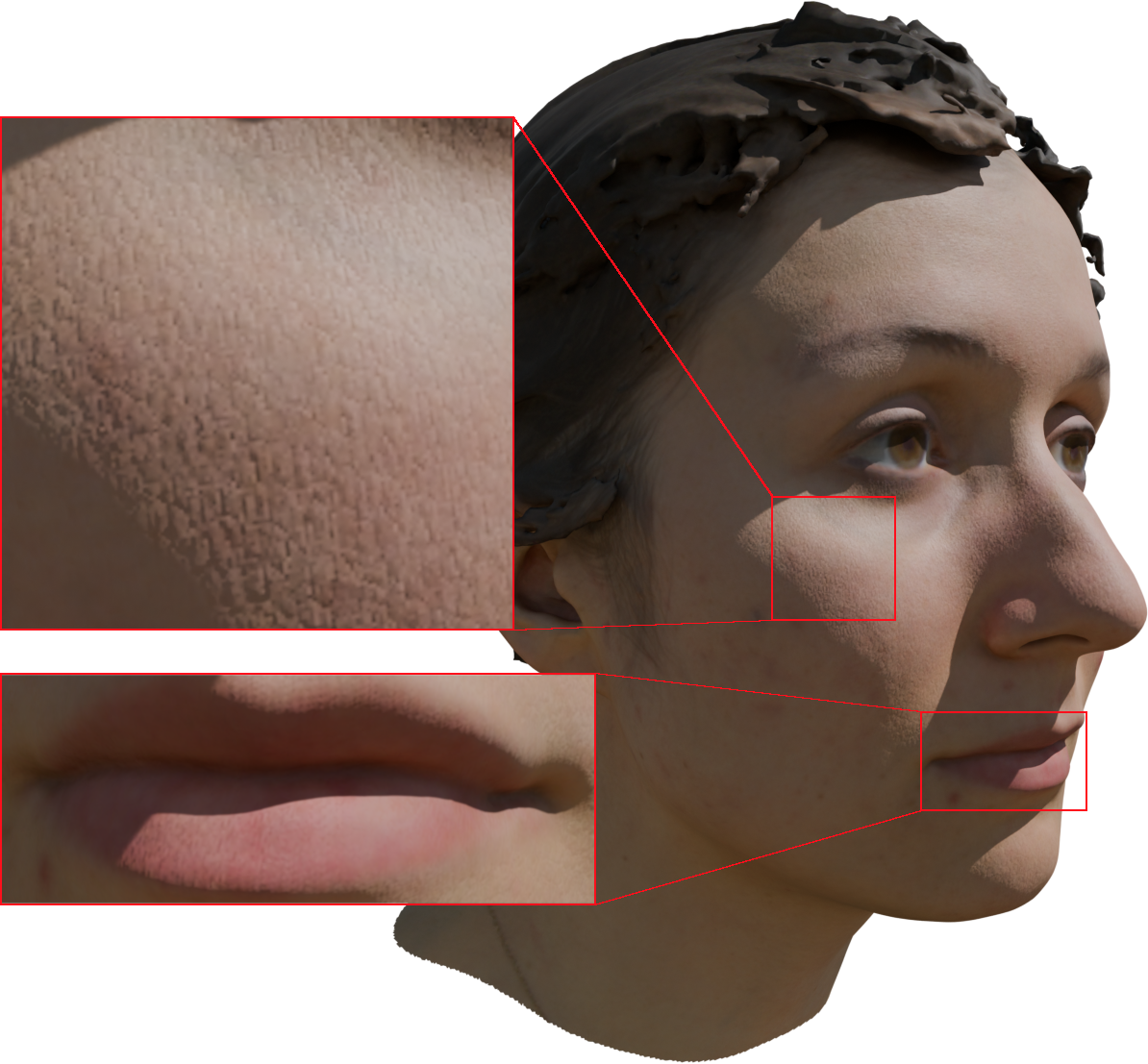}
    \end{subfigure}
    
    \caption{Visual variance of our procedural PBR augmentation pipeline. \textbf{Top Row:} Generated roughness maps showcasing varying simulated skin types, ranging from very dry (left) to oily (right). \textbf{Bottom Row:} Generated tangent-space normal maps demonstrating the effect of our randomised factors $\alpha$ and $\beta$. Ranging from the lowest possible strength (left), to the mid-point of strength (middle), and the highest possible strength (right). The overlaid insets highlight the procedural multi-scale noise used to synthesise micro-pores and directional lip wrinkles.}
    \label{fig:blender_pipeline}
\end{figure}

To generate the large-scale training data, we developed an automated, physics-based augmentation pipeline. The pipeline takes raw geometry and diffuse albedo maps as input and procedurally synthesises physically plausible roughness and tangent-space normal maps. To ensure anatomical accuracy and material fidelity, the synthesis is specialised across four semantic regions: Skin, Mouth/Lips, Hair, and Eyes. The visual variance achieved across all these regions simultaneously---ranging from dry to oily global roughness and young to old structural details---is illustrated in \cref{fig:blender_pipeline}.

\cref{tab:augmentation_params} summarises the key parameters and sampling
distributions for all four regions. Full details for each region are given in the
following subsections.

\begin{table}[tb]
\centering
\caption{Per-region augmentation parameters and sampling distributions. All continuous parameters are sampled uniformly within the stated range unless
otherwise noted.}
\label{tab:augmentation_params}
\resizebox{\linewidth}{!}{%
\begin{tabular}{llll}
\toprule
\textbf{Region} & \textbf{Parameter} & \textbf{Sampling} & \textbf{Range / Values} \\
\midrule
\multirow{4}{*}{Skin}
  & Age Factor $\alpha$       & Uniform    & $[0,\,1]$ \\
  & Skin Type                 & Categorical & 5 classes \\
  & Roughness baseline shift  & Uniform (per class) & $[0.0,\,1.0]$ \\
  & Normal map strength       & Derived from $\alpha$ & $[0.25,\,1.6]$ \\
\midrule
\multirow{3}{*}{Mouth/Lips}
  & Lip fissure strength $\beta$ & Uniform & $[0.15,\,0.5]$ \\
  & Wetness Factor            & Uniform    & $[0.0,\,1.0]$ \\
  & Base roughness            & Fixed offset (lower than skin) & -- \\
\midrule
\multirow{2}{*}{Hair}
  & Oiliness Factor           & Uniform    & $[0.0,\,1.0]$ \\
  & Roughness range           & Derived from oiliness & $[0.55,\,0.80]$ \\
\midrule
\multirow{2}{*}{Eyes}
  & Roughness                 & Fixed      & $0.1$ \\
  & Index of Refraction (IOR) & Fixed      & $1.337$ \\
\bottomrule
\end{tabular}}
\end{table}

\subsection{Skin Material Synthesis}
The skin material is the most complex component of our procedural pipeline, designed to simulate both micro-scale and macro-scale human skin features. 

\paragraph{Normal Maps:} We construct a multi-layered procedural normal map controlled by a continuously sampled \textit{Age Factor} $\alpha \in [0, 1]$. The map consists of three distinct frequencies:
\begin{itemize}
    \item \textbf{Micro Detail (Pores):} Generated using high-frequency Perlin noise \cite{PerlinNoise} (scale 350) to simulate skin pores.
    \item \textbf{Mid Detail (Wrinkles):} Synthesized via Voronoi \cite{Voronoi} distance-to-edge textures. The mapping coordinates are stretched asymmetrically to mimic the directional flow of facial wrinkles.
    \item \textbf{Macro Detail (Unevenness):} Low-frequency Perlin noise (scale 5.0) is added to introduce soft, large-scale surface variations, breaking up artificial smoothness.
\end{itemize}
\Cref{alg:skin_mat} details how the three distinct frequencies are mixed and generated to produce the normal augmentation.

This physically simulates the deeper structural skin variations and loss of elasticity associated with ageing (highlighted in the zoomed-in skin crops in \cref{fig:blender_pipeline}).

\begin{algorithm}[tb]
\caption{Skin Material Synthesis}
\label{alg:skin_mat}
\begin{algorithmic}[1]
\Require $\text{Albedo}_{RGB}$, $\text{Age} \in [0, 1]$, $\text{SkinType} \in [0, 1]$, $\text{Geometry.Pointiness}$, $\text{AmbientOcclusion}$
\Statex
\State \textbf{Normal Map Generation:}
\State $w_{\text{wrinkle}} \gets \text{MapRange}(\text{Age}, [0, 1], [0.05, 0.75])$
\State $s_{\text{bump}} \gets \text{MapRange}(\text{Age}, [0, 1], [0.25, 1.60])$
\State $N_{\text{pores}} \gets \text{NoiseTexture}(\text{scale}=350, \text{detail}=3.0)$
\State $N_{\text{macro}} \gets \text{NoiseTexture}(\text{scale}=5.0, \text{detail}=0.0, \text{roughness}=0.4)$
\State $C_{\text{stretch}} \gets \text{TexCoords} \times (1.5, 3.0, 1.0)$ \Comment{Asymmetric stretch}
\State $V_{\text{wrinkles}} \gets \text{Voronoi}(C_{\text{stretch}}, \text{type}=\text{DistanceToEdge}, \text{scale}=1300)$
\State $M_1 \gets \text{Mix}(N_{\text{pores}}, N_{\text{macro}}, \text{factor}=0.45)$
\State $H_{\text{bump}} \gets \text{Mix}(M_1, V_{\text{wrinkles}}, \text{factor}=w_{\text{wrinkle}})$
\State $\text{Normal}_{\text{out}} \gets \text{Bump}(\text{height}=H_{\text{bump}}, \text{strength}=s_{\text{bump}})$
\Statex
\State \textbf{Roughness Map Generation:}
\State $P \gets \text{ColorRamp}(\text{Geometry.Pointiness}, \text{min}=0.2, \text{max}=0.8)$
\State $M_{\text{geo}} \gets \text{Multiply}(P, \text{AmbientOcclusion}, \text{factor}=0.7)$
\State $V_{\text{inv}} \gets \text{MapRange}(\text{HSV}(\text{Albedo}_{RGB}).V, [0, 1], [1.0, 0.0])$
\State $R_{\text{base}} \gets \text{Mix}(M_{\text{geo}}, V_{\text{inv}}, \text{factor}=0.4)$
\State $R_{\text{target}} \gets \text{MapRange}(\text{SkinType}, [0, 1], [0.65, 0.25])$
\State $\text{Roughness}_{\text{out}} \gets \text{Mix}(R_{\text{base}}, R_{\text{target}}, \text{factor}=0.65)$
\Statex
\State \Return $\text{BSDF}(\text{Color}=\text{Albedo}_{RGB}, \text{Roughness}=\text{Roughness}_{\text{out}}, \text{Normal}=\text{Normal}_{\text{out}}, \text{IOR}=1.44)$
\end{algorithmic}
\end{algorithm}

\paragraph{Roughness Maps:} 
Skin roughness is computed dynamically by blending geometric and colour cues.
We extract geometric \textit{Pointiness} (curvature) and \textit{Ambient Occlusion} (AO) to vary roughness between crevices and exposed areas. Together with the HSV value channel of the albedo map, we generate plausible roughness maps as stated in \cref{alg:skin_mat}.
The input \textit{Skin Type} parameter is uniformly sampled from the range defined by the draw of one of the five categories, which have predefined probabilities:

\begin{itemize}
    \item \textbf{Very Dry} (10\%): roughness $\in [0.0,\,0.15]$
    \item \textbf{Dry} (17.75\%): roughness $\in [0.15,\,0.38]$
    \item \textbf{Normal} (40\%): roughness $\in [0.38,\,0.62]$
    \item \textbf{Combination} (20\%): roughness $\in [0.62,\,0.85]$
    \item \textbf{Oily} (12.25\%): roughness $\in [0.85,\,1.0]$
\end{itemize}

\subsection{Specialized Semantic Regions}

\paragraph{Mouth / Lips:} The lip region requires distinct structural details. We lower the base roughness compared to the surrounding skin and generate a specialised normal map, detailed in \cref{alg:lips_mat}. We simulate the characteristic vertical fissuring of human lips, where the strength is determined by a uniformly sampled factor $\beta \in [0.15,0.5]$ (visible in the lip insets of \cref{fig:blender_pipeline}). Additionally, a randomised \textit{Wetness Factor} controls the final roughness, allowing the pipeline to generate both matte, dry, and highly wet lips.
\paragraph{Hair:} To simulate hair, we sample a random \textit{Oiliness Factor}. This drives a mapped roughness range, producing variations from dry, highly scattering hair (roughness 0.8) to darker, oily, and clumped hair (roughness 0.55), as outlined in \cref{alg:hair_eyes_mat}. We deliberately omit normal map augmentation for the hair region. Procedurally generating high-quality, flow-aligned normal maps for diverse hair styles and topologies---relying solely on coarse geometry and albedo---is highly ill-posed. In our experiments, applying synthetic normals to hair introduced structurally inconsistent artefacts that degraded rather than improved rendering fidelity.
\paragraph{Eyes:} The eyeballs are assigned a static, physically motivated material with a fixed low roughness (0.1) and an Index of Refraction (IOR) of 1.337, as specified in \cref{alg:hair_eyes_mat}, matching the physical properties of the human cornea to simulate a wet, glassy appearance. Normal map augmentation is entirely bypassed for the eyes, as the natural surface of the cornea is exceptionally smooth and procedural normals would introduce spurious micro-structure inconsistent with the ground-truth surface.
\begin{algorithm}[tb]
\caption{Lips Material Synthesis}
\label{alg:lips_mat}
\begin{algorithmic}[1]
\Require $\text{Albedo}_{RGB}$, $\text{Wetness} \in [0, 1]$
\Statex
\State \textbf{Normal Map Generation:}
\State $S \gets \text{RandomUniform}(0.15, 0.50)$
\State $C_{\text{stretch}} \gets \text{TexCoords} \times (25.0, 1.0, 8.0)$ \Comment{Extreme vertical stretch}
\State $N_{\text{lines}} \gets \text{NoiseTexture}(\text{coords}=C_{\text{stretch}}, \text{scale}=100.0)$
\State $\text{Normal}_{\text{out}} \gets \text{Bump}(\text{height}=N_{\text{lines}}, \text{strength}=S)$
\Statex
\State \textbf{Roughness Map Generation:}
\State $\text{Roughness}_{\text{out}} \gets \text{MapRange}(\text{Wetness}, [0, 1], [0.65, 0.30])$
\Statex
\State \Return $\text{BSDF}(\text{Color}=\text{Albedo}_{RGB}, \text{Roughness}=\text{Roughness}_{\text{out}}, \text{Normal}=\text{Normal}_{\text{out}}, \text{IOR}=1.44, \text{Clearcoat}=0.05)$
\end{algorithmic}
\end{algorithm}

\begin{algorithm}[tb]
\caption{Hair and Eyes Material Synthesis}
\label{alg:hair_eyes_mat}
\begin{algorithmic}[1]
\Require $\text{Albedo}_{RGB}$, $\text{Oiliness} \in [0, 1]$
\Statex
\State \textbf{Hair Material Generation:}
\State $R_{\text{base}} \gets \text{MapRange}(\text{Oiliness}, [0, 1], [0.80, 0.55])$
\State $V_{\text{albedo}} \gets \text{HSV}(\text{Albedo}_{RGB}).V$
\State $\Delta_R \gets \text{MapRange}(V_{\text{albedo}}, [0, 1], [0.05, -0.05])$ \Comment{Slight perturbation based on HSV value}
\State $\text{Roughness}_{\text{hair}} \gets \text{Clamp}(R_{\text{base}} + \Delta_R, 0, 1)$
\State $\text{BSDF}_{\text{hair}} \gets \text{BSDF}(\text{Color}=\text{Albedo}_{RGB}, \text{Roughness}=\text{Roughness}_{\text{hair}}, \text{IOR}=1.55)$
\Statex
\State \textbf{Eyes Material Generation:}
\State $\text{BSDF}_{\text{eyes}} \gets \text{BSDF}(\text{Color}=\text{Albedo}_{RGB}, \text{Roughness}=0.10, \text{IOR}=1.337)$
\end{algorithmic}
\end{algorithm}

\section{Architectural Details} 
\label{supp:ArchitecturalDetails}

The high-level network configurations, training setup, and data generation
procedure are described in Secs.~3.1.--3.5. of the main paper. This section
provides the fine-grained architectural details deferred from the main paper
due to space constraints: 
the full CNN architectures (\cref{sec:cnn_architectures}), a consolidated hyperparameter table (\cref{tab:hyperparams}), and the complete noise scheduling and inference
sampling specification (\cref{sec:noise_scheduling}).

\begin{table}[t]
\centering
\caption{Architectural and noise-scheduling hyperparameters for the albedo
and roughness prediction networks.
}
\label{tab:hyperparams}
\begin{tabular}{lll}
\toprule
\textbf{Hyperparameter} & \textbf{Albedo} & \textbf{Roughness} \\
\midrule
Base feature dimensionality $d$   & 64                  & 64 \\
Attention heads                   & 4                   & 4 \\
Cross-attention patch size $r$    & \textit{5}  & \textit{5} \\
KNN neighbourhood size $k$        & \textit{21}  & -- (omitted) \\
EDBs with cross-attention         & \{1, 4, 5, 6, 7, 9\} & \{1, 2, 3, 4\} \\
EDBs with $k$-NN attention        & \{1, 4, 5, 6, 7, 9\} & -- (omitted) \\
LBO eigenpairs $K$                & 300                 & 300 \\
Input conditioning space          & RGB $[-1, 1]$       & HSV \\
Batch size                        & 32                  & 32 \\
Warm-up steps                           & 2000 steps          & 2000 steps \\
Learning rate (lr)                     & $10^{-4}$           & $10^{-4}$ \\
lr-scheduler                & cosine & cosine \\
Optimiser                         & AdamW~\cite{loshchilov2017decoupled} & AdamW~\cite{loshchilov2017decoupled} \\
\midrule
$P_{\text{mean}}$ (high-noise)    & $-0.15$             & $-0.15$ \\
$P_{\text{std}}$ (high-noise)     & $1.00$              & $1.00$ \\
$P_{\text{mean}}$ (low-noise)     & $-2.45$             & $-2.45$ \\
$P_{\text{std}}$ (low-noise)      & $0.95$              & $0.95$ \\
$\sigma_{\text{data}}$            & $0.391$             & $0.391$ \\
Hand-off threshold $t_i$          & $\approx 0.55$      & $\approx 0.55$ \\
Inference steps $N$               & 31                  & 31 \\
$\sigma_{\text{max}}$             & 80                  & 80 \\
$\sigma_{\text{min}}$             & 0.002               & 0.002 \\
$\rho$                            & 7                   & 7 \\
\bottomrule
\end{tabular}
\end{table}

\subsection{CNN Architectures}
\label{sec:cnn_architectures}

Image conditioning and tangent-space refinement rely on two specialised
CNNs. Their architectures are detailed below:
\paragraph{Image Pre-Processing CNN.}
This network maps input images into a multi-scale feature space and
progressively downsamples them, enabling the Image-to-Mesh Cross-Attention
to operate over varying receptive fields at each EDB (\cref{tab:cnn_image}).

\begin{table}[t]
\centering
\caption{Architecture of the Image Pre-Processing CNN. The mapping block
projects input images to feature space, followed by three sequential
downsampling blocks.}
\label{tab:cnn_image}
\resizebox{\linewidth}{!}{%
\begin{tabular}{llllll}
\toprule
\textbf{Block} & \textbf{Layer Type} & \textbf{Kernel / Stride}
    & \textbf{Out Channels} & \textbf{Activation, Norm} \\
\midrule
\multirow{2}{*}{Mapping}
    & Conv2d & $3{\times}3$ / 1 & 16 & GroupNorm(2), SiLU \\
    & Conv2d & $3{\times}3$ / 1 & 32 & GroupNorm(4), SiLU \\
Downsample 1 & Conv2d & $3{\times}3$ / 2 & 32 & GroupNorm(4), SiLU \\
Downsample 2 & Conv2d & $3{\times}3$ / 2 & 32 & GroupNorm(4), SiLU \\
Downsample 3 & Conv2d & $3{\times}3$ / 2 & 32 & GroupNorm(4), SiLU \\
\bottomrule
\end{tabular}
}
\end{table}
\paragraph{Tangent Refinement U-Net.}
For the eye regions, $64{\times}64$ patches extracted in tangent space are
processed by a lightweight U-Net to recover sharp, high-frequency details
such as iris structure. The final projection layer uses zero-initialisation
to ensure stable residual learning at the start of training, fully shown in \cref{tab:unet}.

\begin{table}[t]
\centering
\caption{Architecture of the Tangent Refinement U-Net. This lightweight
network processes $64{\times}64$ tangent-space patches to recover
high-frequency eye and iris details.}
\label{tab:unet}
\resizebox{\linewidth}{!}{%
\begin{tabular}{lllll}
\toprule
\textbf{Block} & \textbf{Layer Type} & \textbf{Kernel / Stride}
    & \textbf{Out Channels} & \textbf{Activation, Norm} \\
\midrule
Encoder 1   & Conv2d $\times$2    & $3{\times}3$ / 1 & 32  & GroupNorm(8), SiLU \\
Down 1      & Conv2d              & $3{\times}3$ / 2 & 64  & -- \\
Encoder 2   & Conv2d              & $3{\times}3$ / 1 & 64  & GroupNorm(16), SiLU \\
Down 2      & Conv2d              & $3{\times}3$ / 2 & 128 & -- \\
Bottleneck  & Conv2d              & $3{\times}3$ / 1 & 128 & GroupNorm(32), SiLU \\
Up 1        & ConvTranspose2d     & $2{\times}2$ / 2 & 64  & -- \\
Concat 1    & Concat w/ Encoder 2 & --               & 128 & -- \\
Decoder 1   & Conv2d              & $3{\times}3$ / 1 & 64  & GroupNorm(16), SiLU \\
Up 2        & ConvTranspose2d     & $2{\times}2$ / 2 & 32  & -- \\
Concat 2    & Concat w/ Encoder 1 & --               & 64  & -- \\
Decoder 2   & Conv2d              & $3{\times}3$ / 1 & 32  & GroupNorm(8), SiLU \\
Final       & Conv2d (Zero Init)  & $1{\times}1$ / 1 & $c_{\text{out}}$ & -- \\
\bottomrule
\end{tabular}}
\end{table}

\subsection{Noise Scheduling and Inference}
\label{sec:noise_scheduling}

Following the EDM framework~\cite{EDM}, we train two specialised models to handle distinct noise regimes for both albedo and roughness prediction. The injected noise during
training is:
\begin{equation}
    \text{noise} = \mathbf{n} \cdot \sigma, \quad
    \mathbf{n} \sim \mathcal{N}(\mathbf{0}, \mathbf{I}),
\end{equation}
where the noise level $\sigma$ is drawn from a log-normal distribution:
\begin{equation}
    \sigma = \exp\!\left(r \cdot P_{\text{std}} + P_{\text{mean}}\right),
    \quad r \sim \mathcal{N}(0, 1).
\end{equation}
The noise regimes are divided into high-noise and low-noise regimes, detailed in \cref{tab:hyperparams}.

\paragraph{Loss Weighting.}
We use the standard EDM loss weighting. The per-sample weight for noise
level $\sigma$ is:
\begin{equation}
    w(\sigma) = \frac{\sigma^2 + \sigma_{\text{data}}^2}
                     {(\sigma \cdot \sigma_{\text{data}})^2},
\end{equation}
with $\sigma_{\text{data}} = 0.391$ computed as the standard deviation
of our training data.

\paragraph{Inference Sampling.}
During inference, we denoise over $N = 31$ steps with non-uniform time
step discretisation concentrating steps at lower noise levels:
\begin{equation}
    t_i = \left(
        \sigma_{\text{max}}^{1/\rho} +
        \frac{i}{N-1}
        \left(
            \sigma_{\text{min}}^{1/\rho} - \sigma_{\text{max}}^{1/\rho}
        \right)
    \right)^{\!\rho},
    \quad i \in \{0, \ldots, 30\},
\end{equation}
with $\sigma_{\text{max}} = 80$, $\sigma_{\text{min}} = 0.002$,
and $\rho = 7$.

\paragraph{Two-Stage Hand-Off.}
We perform a hard hand-off from the high-noise base model to the
low-noise refinement model at $t_i \approx 0.55$, corresponding to
the final ${\sim}33\%$ of the denoising trajectory. The high-noise
model establishes coarse global structure and colour, while the
low-noise refinement model recovers fine-grained surface detail.

\subsection{Efficiency}
We benchmarked all methods on an NVIDIA RTX 3090 GPU with 12 CPU cores (\cref{tab:Efficiency}). Our core generative network (7.22M parameters) is $370\times$ smaller than Hunyuan3D-2.0-Paint. This makes VISAGE considerably more practical for deployment than large diffusion models. Through semantic region splitting, our full model achieves a lower peak memory footprint (5.6\,GB) than even the base DiffusionNet* (6.4\,GB).  Our feed-forward approach synthesises textures in an end-to-end runtime of 104.9\,s, drastically outperforming optimisation-based methods like MonoNPHM (575.4\,s total).
\begin{table}[htbp]
    \centering
    \caption{\textbf{Efficiency comparison.} Parameter counts (M) and inference times (seconds) are split into the core generative network and the off-the-shelf preprocessing components (if applicable)}
        \adjustbox{width=\columnwidth}{%
    \begin{tabular}{lccc}
        \toprule
        \textbf{Model} & \textbf{Parameter Count}  & \textbf{GPU Memory} & \textbf{Inference Time} \\
        \midrule
        DiffusionNet* \cite{DifNet} & 3.51M + 214.79M & \phantom{0}6.40GB & 21.3\,s + 78.0\,s \\
        Hunyuan3D-2.0-Paint \cite{Hunyuan3d} & 2679.13M & 10.30GB & 87.9\,s  \\
        MonoNPHM \cite{MonoNPHM}& 11.22M + 84.79M & \phantom{0}4.96GB & 417.5\,s + 157.9\,s \\
        VISAGE & 7.22M + 214.79M & \phantom{0}5.60GB & 26.9\,s + 78.0\,s  \\
        \bottomrule
    \end{tabular}
    }
    \label{tab:Efficiency}
\end{table}

\subsection{Choice of Spectral Basis Size}
\begin{figure}[htbp]
    \centering
    \includegraphics[width=1\columnwidth]{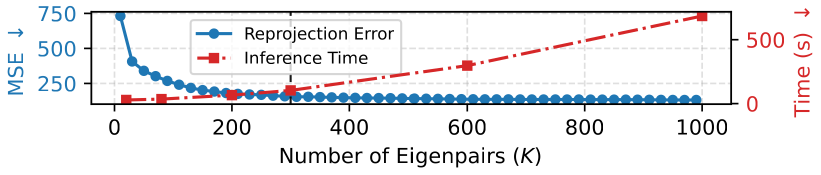}
    \caption{\textbf{Spectral reconstruction trade-off.} Increasing the number of Laplace--Beltrami eigenpairs reduces reprojection error but increases inference time. We use $K=300$ as a practical trade-off between accuracy and efficiency.}
    \label{fig:eigen_tradeoff}
\end{figure}
\label{supp:eigen}
The number of eigenpairs is chosen by analysing the trade-off between spectral
reconstruction fidelity and computational cost. As shown in
\cref{fig:eigen_tradeoff}, the reprojection error exhibits clear diminishing
returns beyond $K \approx 300$, while inference cost continues to grow
superlinearly. We therefore use $K = 300$ throughout.

\section{Additional Ablations and Visualizations} \label{supp:AddAblations}

This section provides visual as well as ablation results complementing Sec.~4 of
the main paper.
\Cref{supp:Noise} quantifies the model robustness to noisy meshes.
\Cref{supp:roughness} isolates the contribution of the predicted roughness maps.
\Cref{supp:visualablation} presents the qualitative ablation comparisons
deferred from Sec.~4.2.1.
\Cref{supp:viz:failures} characterises representative failure cases on the
NPHM and FaceVerse datasets.
\subsection{Robustness to Noise}
\label{supp:Noise}
\begin{figure}
    \centering
    \includegraphics[width=\columnwidth]{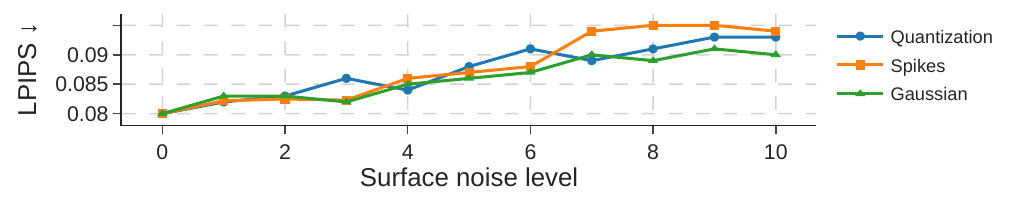}
    \caption{\textbf{Robustness to geometric noise.} LPIPS under Gaussian noise, quantization, and surface spikes. VISAGE degrades gracefully under increasing mesh perturbations, indicating robustness to imperfect input geometry.}
    \small
    \label{fig:Mesh}
\end{figure}

Our pipeline mitigates preprocessing fragility via two safeguards: (1) EPnP+RANSAC rejects outlier keypoints, and (2) training on estimated parameters implicitly conditions the network against estimation jitter. 
To evaluate robustness to geometric artefacts, we systematically perturbed input meshes with three noise variants: (1) scale-invariant Gaussian jitter ($\sigma = 15{-}30\%$ of the mean edge length), (2) structural spikes of increasing density and magnitude, and (3) quantisation noise with expanding step sizes. \cref{fig:Mesh} shows that while the quality suffers with the addition of noise, the impact is marginal even in the case of higher surface noise.

\subsection{Contribution of Predicted Roughness}
\label{supp:roughness}
\begin{figure*}[tbph]
\centering
\setlength{\tabcolsep}{1pt}
\renewcommand{\arraystretch}{0.9}

\begin{tabular}{ccc  }%
    \textbf{GT} & \textbf{Constant $r$} & \textbf{Predicted $r$} \\ %

    \includegraphics[width=0.2\textwidth]{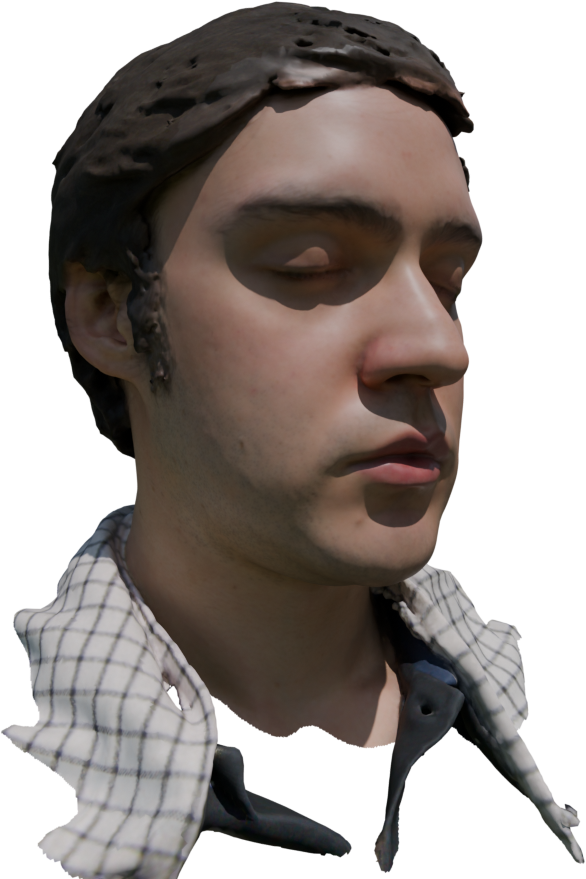} &
    \includegraphics[width=0.2\textwidth]{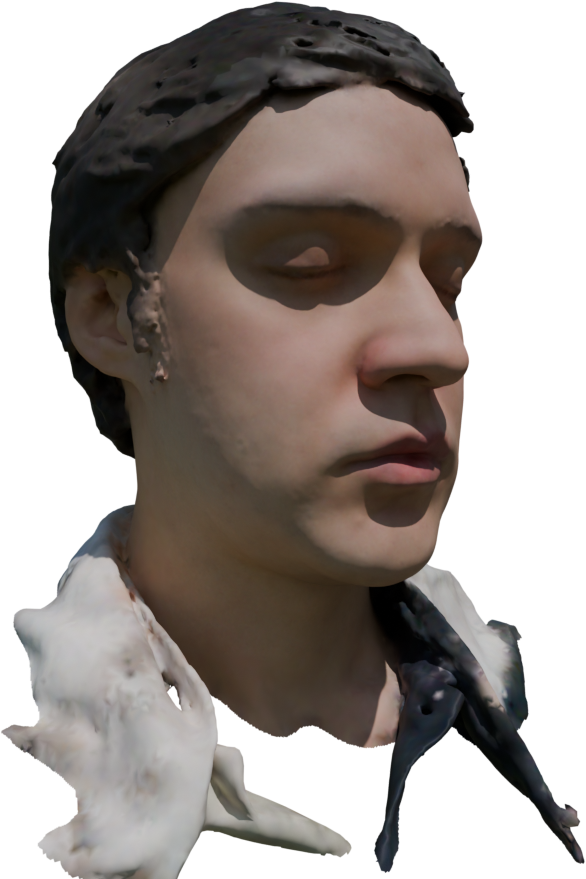} &
    \includegraphics[width=0.2\textwidth]{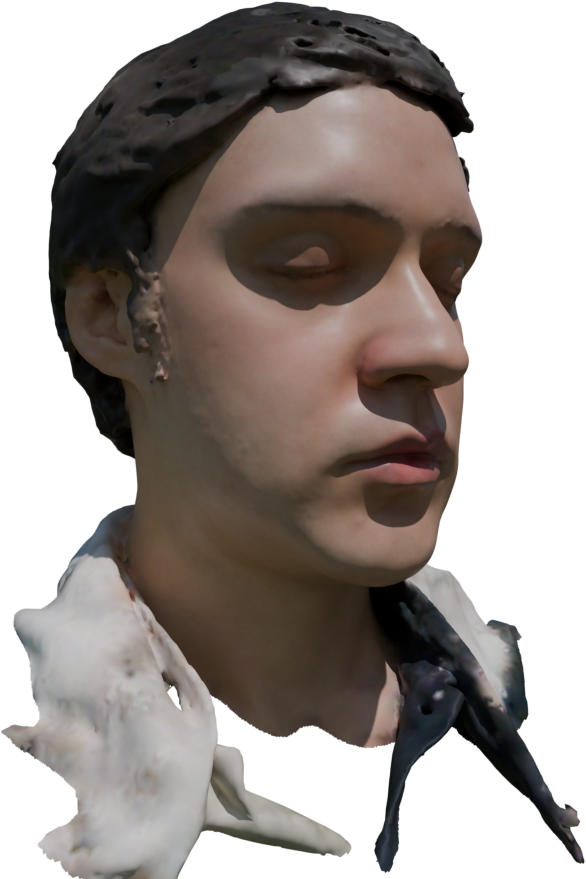} \\
    \includegraphics[width=0.2\textwidth]{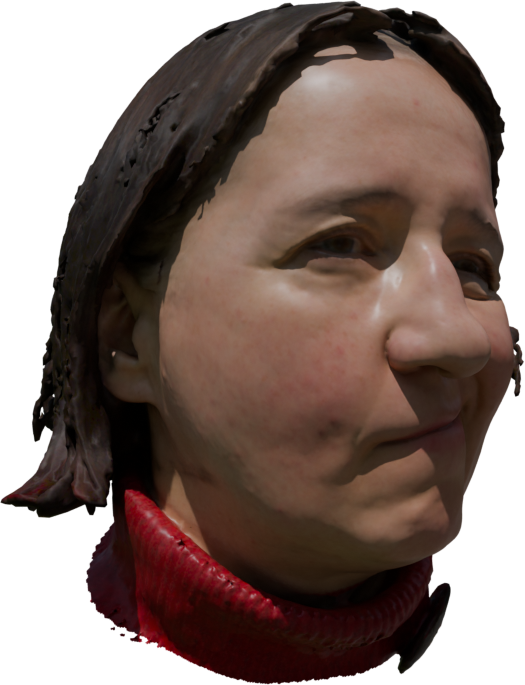} &
    \includegraphics[width=0.2\textwidth]{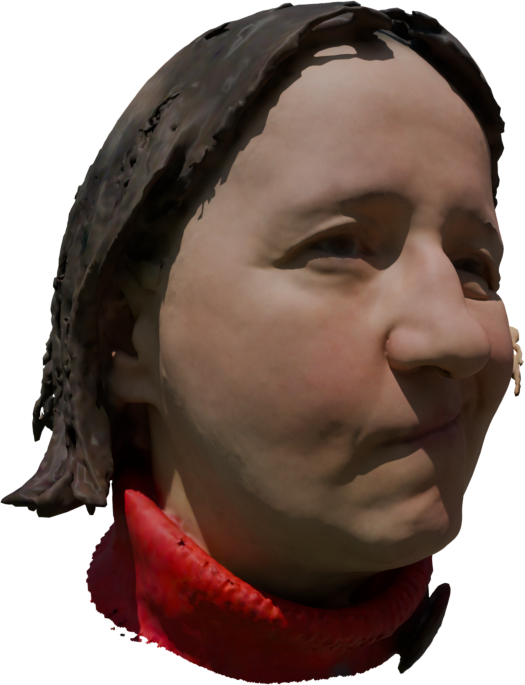} &
    \includegraphics[width=0.2\textwidth]{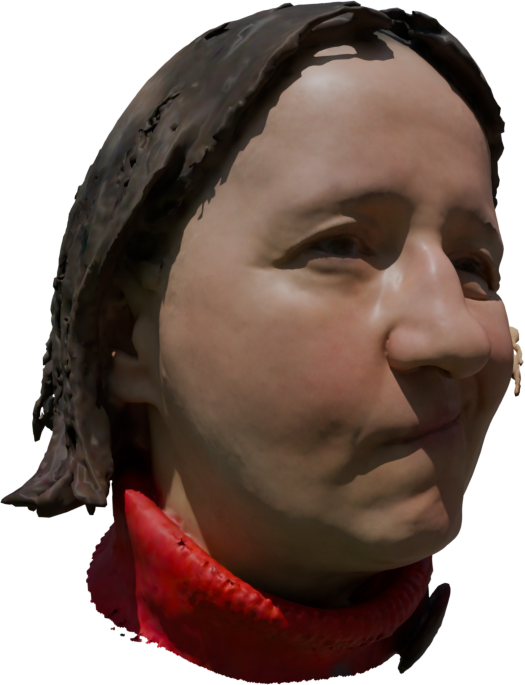} \\

    \includegraphics[width=0.2\textwidth]{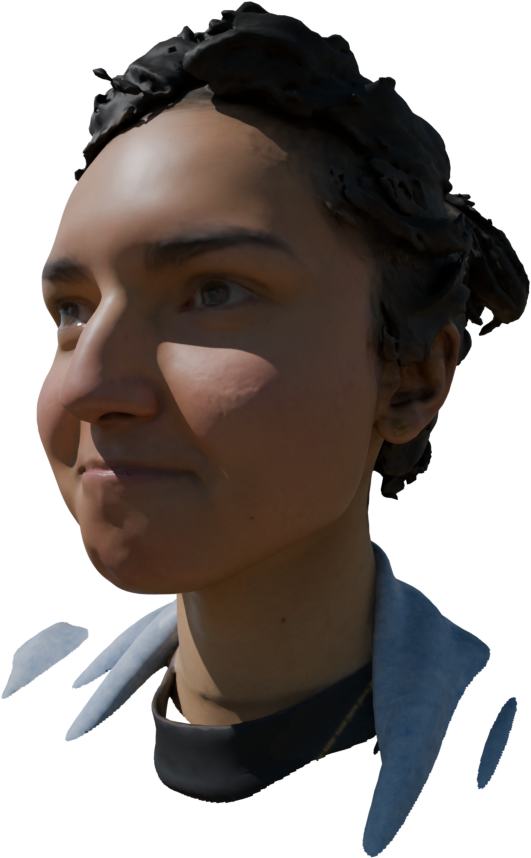} &
    \includegraphics[width=0.2\textwidth]{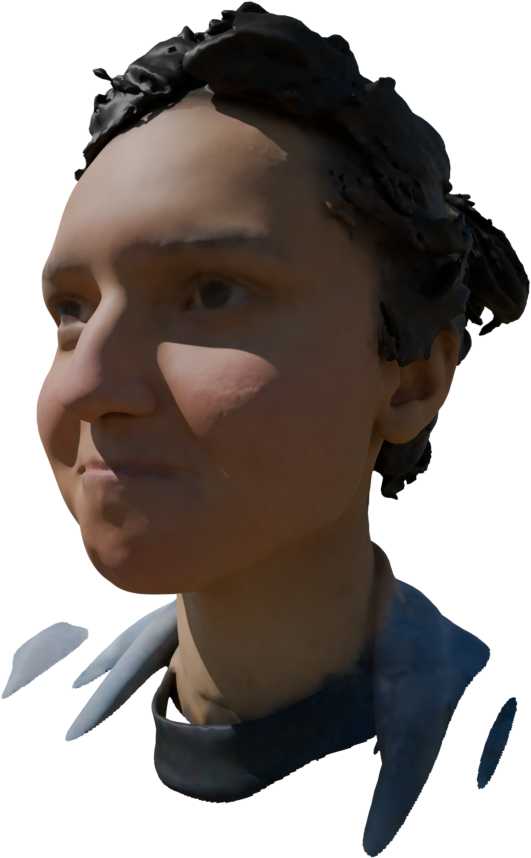} &
    \includegraphics[width=0.2\textwidth]{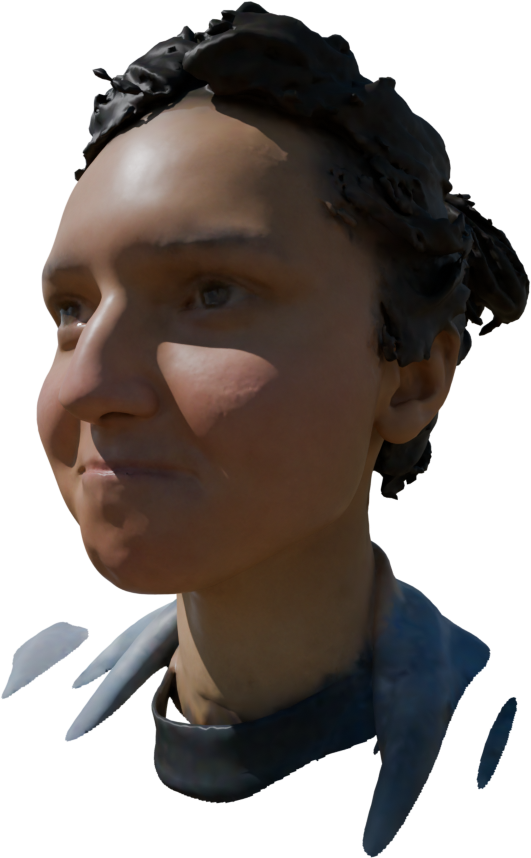}  \\ %
    \includegraphics[width=0.2\textwidth]{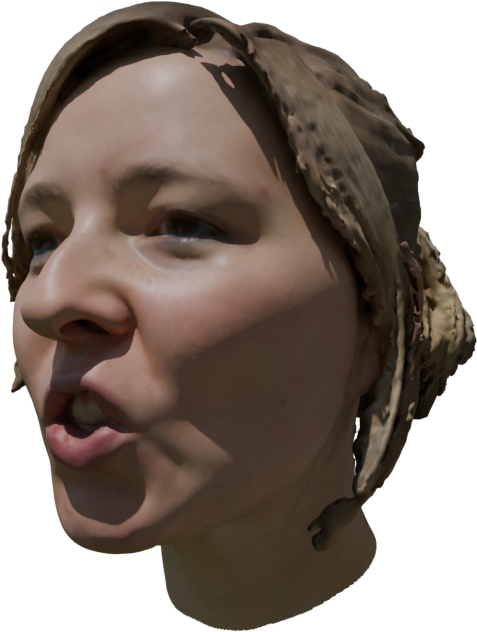} &
    \includegraphics[width=0.2\textwidth]{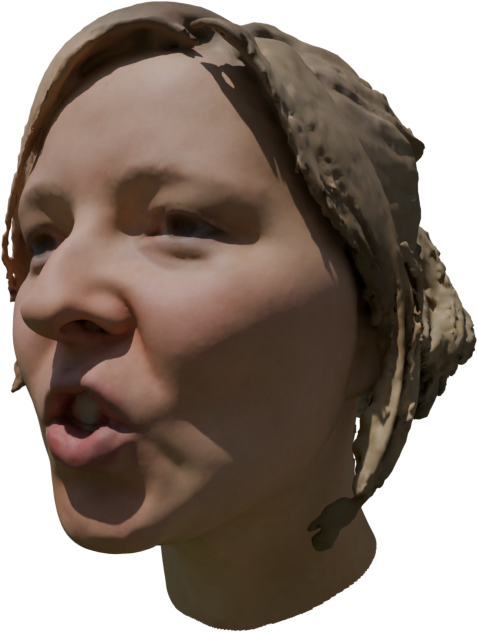} &
    \includegraphics[width=0.2\textwidth]{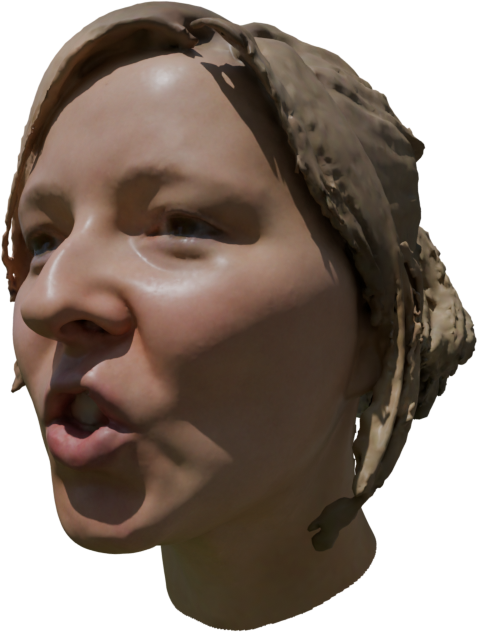} \\
\end{tabular}

\caption{\textbf{Contribution of predicted roughness on NPHM dataset \cite{NPHM}.} Predicted albedo
rendered with a single global constant roughness ($r = 0.59$, the
area-weighted training-set mean) against the predicted per-vertex roughness,
alongside the ground-truth render. All conditions share geometry, albedo,
camera and illumination.}
\label{fig:supp_roughness}
\end{figure*}

To isolate the roughness model's contribution, we render the predicted albedo
under two roughness conditions with geometry, camera and lighting held fixed:
a single global constant equal to the area-weighted training-set mean
($r = 0.59$), which is the obvious fallback for a pipeline without a roughness
model, and our predicted per-vertex roughness. Both are shown alongside the
ground-truth render for four subjects in \cref{fig:supp_roughness}. Predicted roughness
concentrates a response where skin is oiliest --- the forehead and
nose bridge gain a visible sheen, and the eyes read as glossy rather than
matte --- which relieves the flat appearance of the
constant-roughness render. The predicted maps also differ noticeably between
subjects, so the model is not collapsing to a learned constant.

\subsection{Visual Ablation} \label{supp:visualablation}

\begin{figure*}[!htp]
    \centering  
    \begin{subfigure}[b]{0.26\linewidth}
        \includegraphics[width=\linewidth]{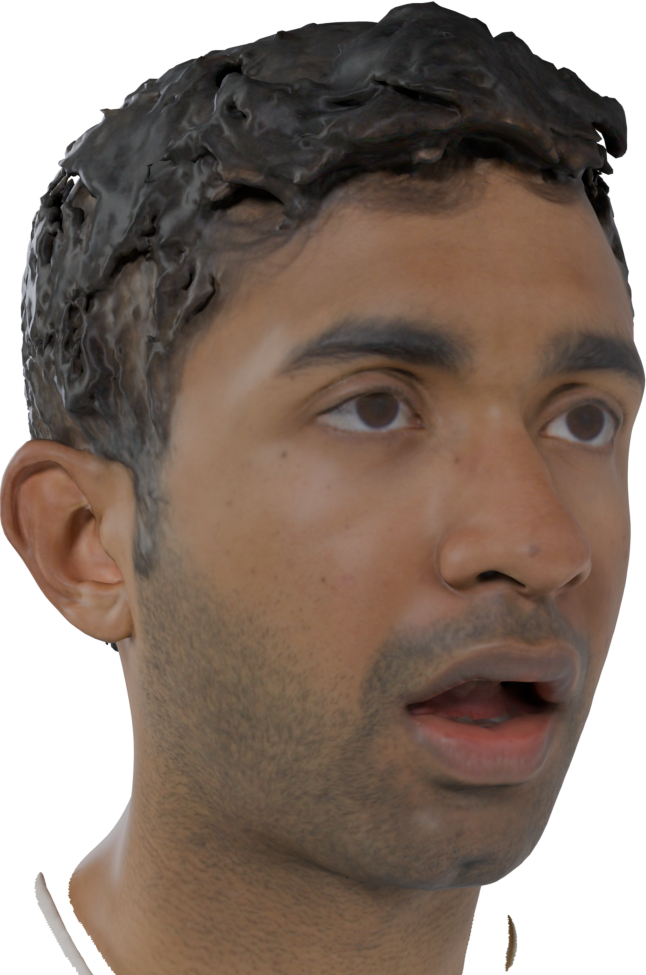}
        \caption*{Ground Truth}
    \end{subfigure}\hfill
    \begin{subfigure}[b]{0.26\linewidth}
        \includegraphics[width=\linewidth]{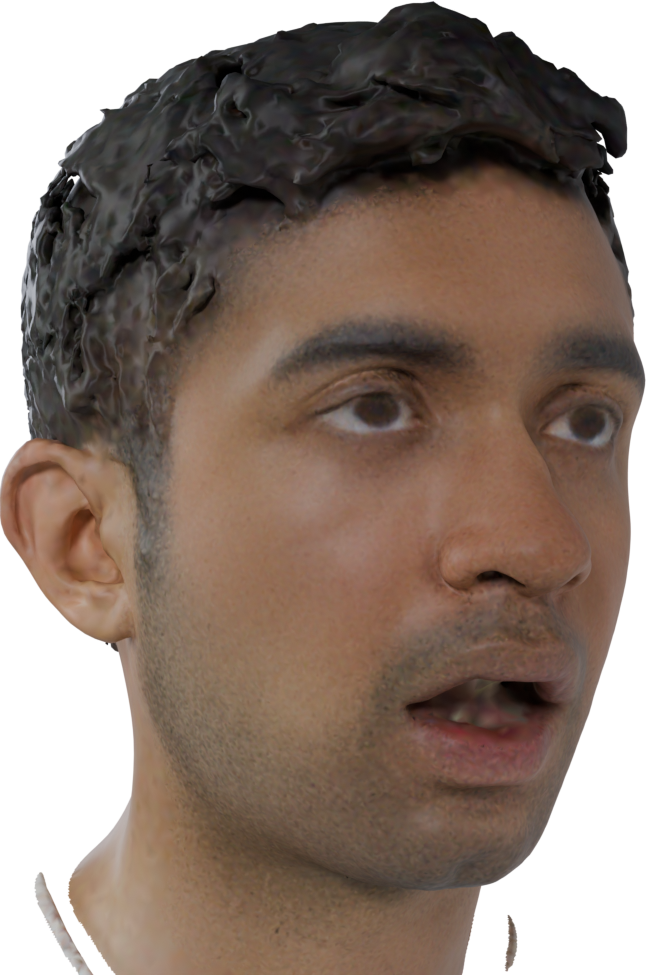}
        \caption*{w/o High-frequency stream}
    \end{subfigure}\hfill
    \begin{subfigure}[b]{0.26\linewidth}
        \includegraphics[width=\linewidth]{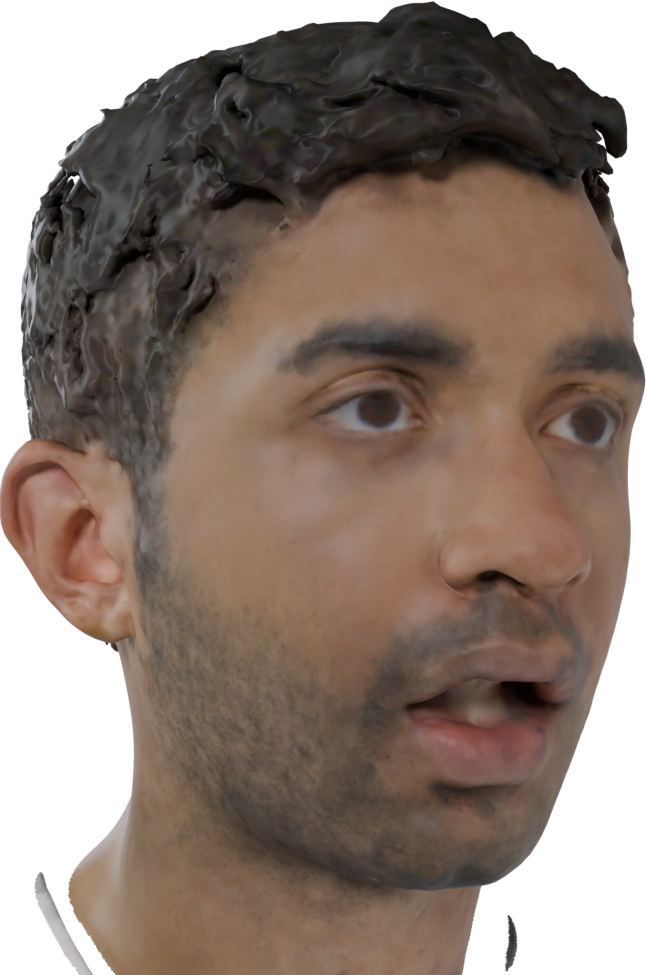}
        \caption*{VISAGE}
    \end{subfigure}
    \vspace{0.1cm} 
    \begin{subfigure}[b]{0.27\linewidth}
        \includegraphics[width=\linewidth]{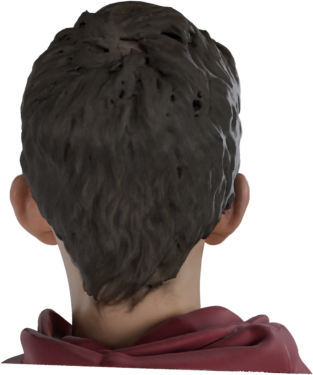}
        \caption*{Ground Truth}
    \end{subfigure}\hfill
    \begin{subfigure}[b]{0.27\linewidth}
        \includegraphics[width=\linewidth]{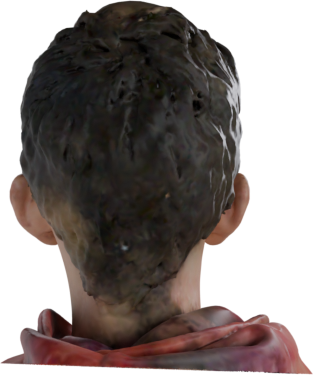}
        \caption*{w/o Noise level conditioning}
    \end{subfigure}\hfill
    \begin{subfigure}[b]{0.27\linewidth}
        \includegraphics[width=\linewidth]{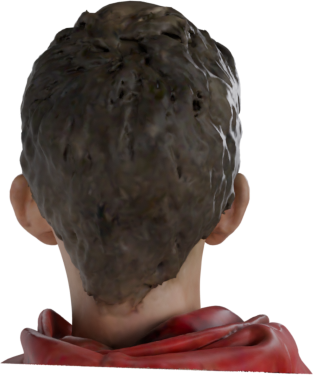}
        \caption*{VISAGE}
    \end{subfigure}
    \vspace{0.1cm}
    \begin{subfigure}[b]{0.24\linewidth}
        \includegraphics[width=\linewidth]{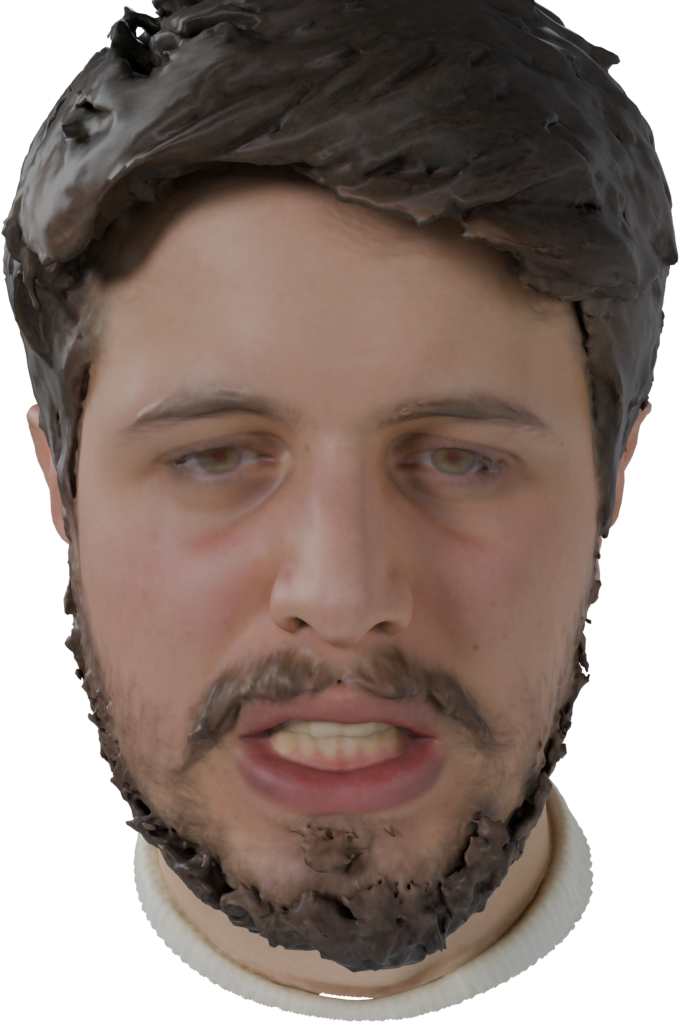}
        \caption*{Ground Truth}
    \end{subfigure}\hfill
    \begin{subfigure}[b]{0.24\linewidth}
        \includegraphics[width=\linewidth]{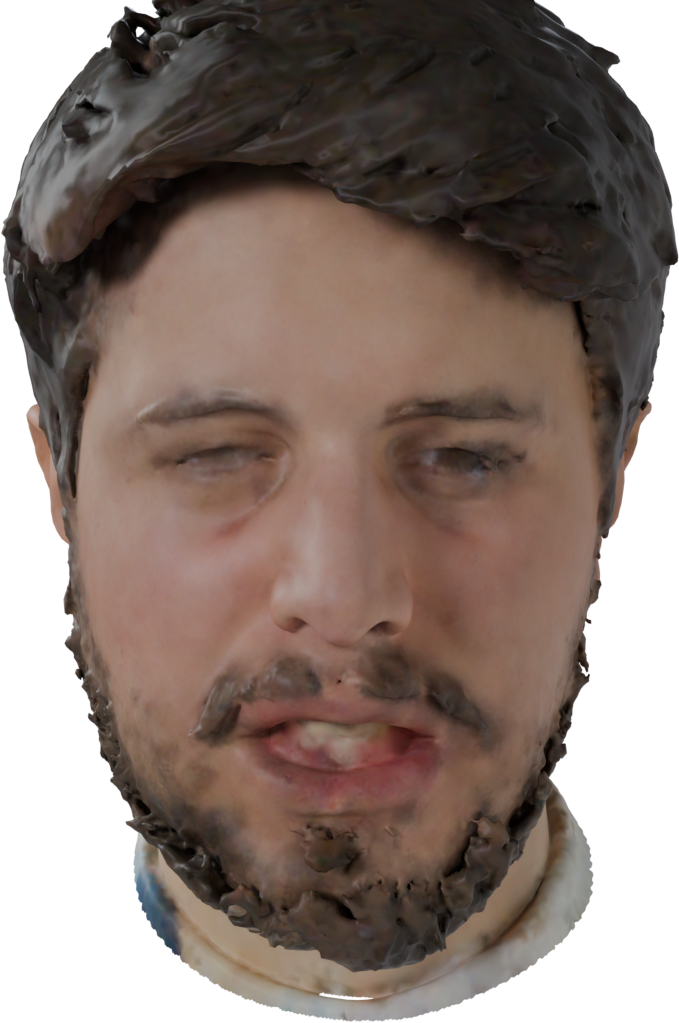}
        \caption*{w/o Region splitting}
    \end{subfigure}\hfill
    \begin{subfigure}[b]{0.24\linewidth}
        \includegraphics[width=\linewidth]{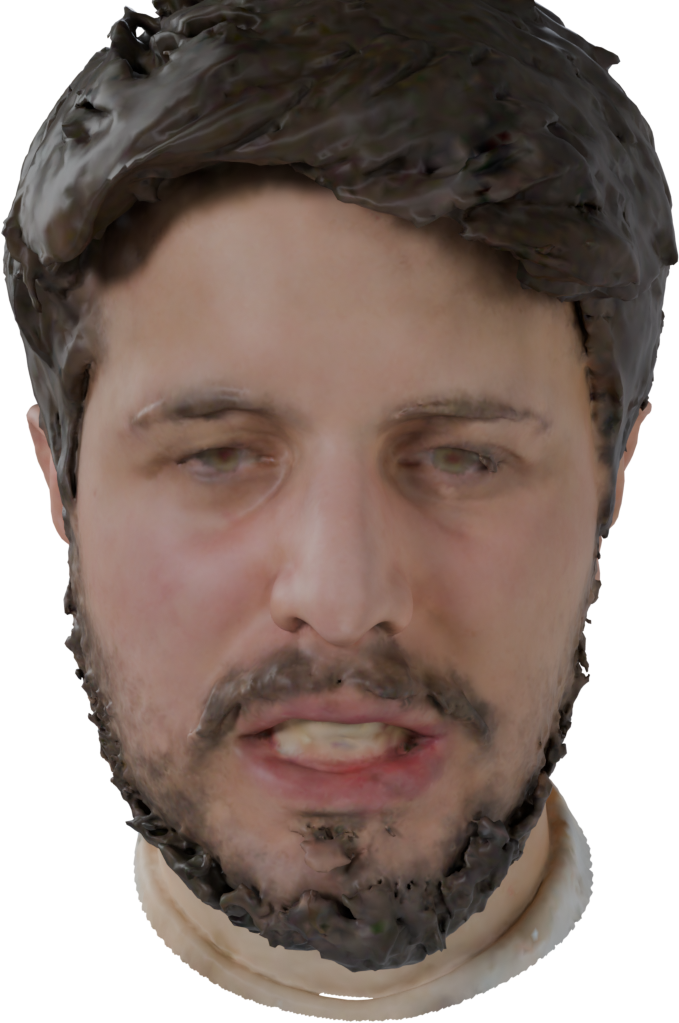}
        \caption*{VISAGE}
    \end{subfigure}
    \caption{Qualitative ablation of key architectural components on the NPHM dataset. \textbf{Top Row:} Removing the local $k$-NN refinement stream forces the network to rely on the EDB's MLP, resulting in grainy artefacts and a loss of high-frequency sharpness, like beard. \textbf{Middle Row:} Disabling noise-level conditioning in the heat diffusion step prevents dynamic spatial adaptation, causing texture bleeding and boundary collapse in occluded areas. \textbf{Bottom Row:} Without semantic region splitting, a shared global MLP fails to capture the unique statistical distributions of specialised regions such as the mouth and eyes, degrading realism.}
    \label{fig:visual_ablation}
\end{figure*}
As described in Sec.~4.2.1 of the main paper, we evaluate four degraded
configurations by systematically removing key architectural components.
The main paper presents qualitative results for the eye region (Fig.~6), as it is the most sensitive indicator of semantic region
splitting and tangent-space refinement.
\cref{fig:visual_ablation} provides the remaining qualitative comparisons for the other three
configurations: removal of the high-frequency $k$-NN stream, removal
of noise-level conditioning, and removal of semantic region splitting.

\begin{figure*}[bhpt]
\centering
\setlength{\tabcolsep}{1pt}
\renewcommand{\arraystretch}{0.9}
\begin{tabular}{ccc}
    \textbf{GT} & \textbf{w/o Stream 2} & \textbf{VISAGE} \\
    \includegraphics[width=0.3\textwidth]{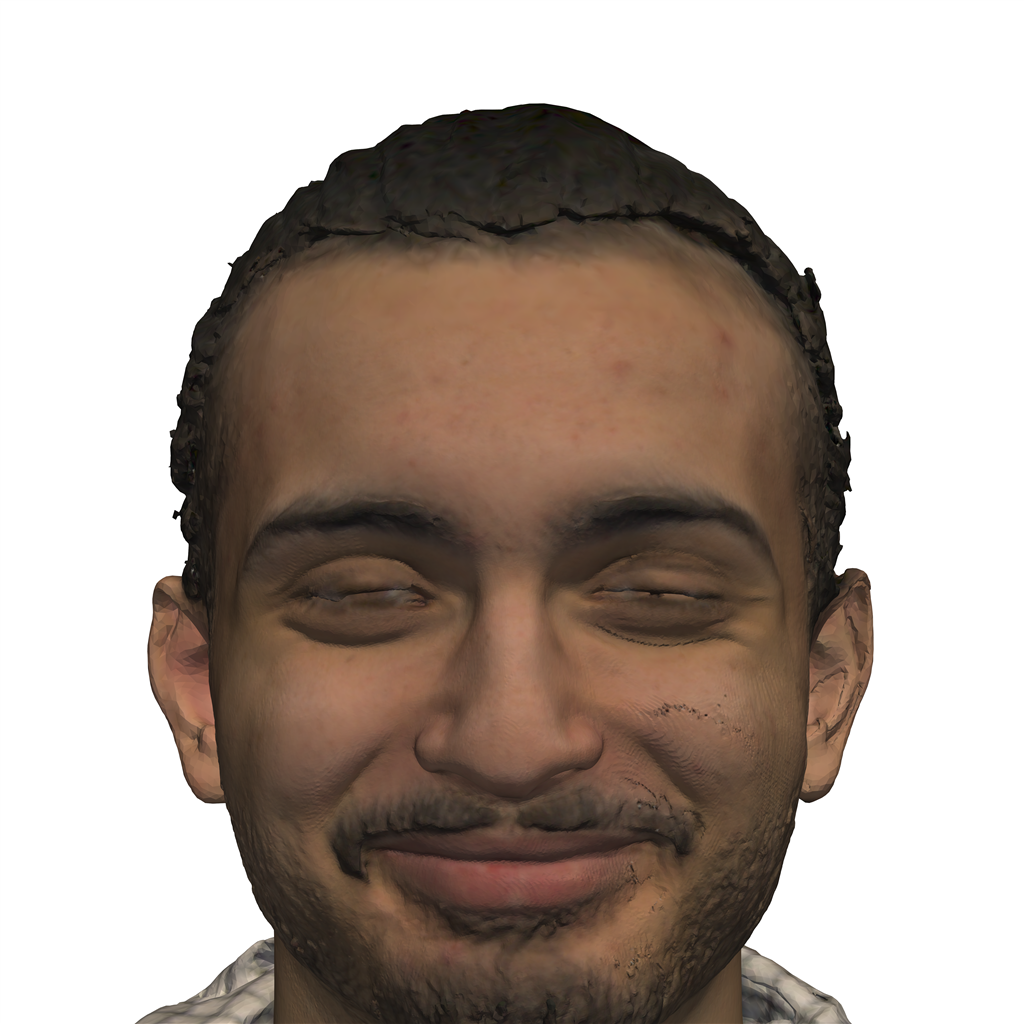} &
    \includegraphics[width=0.3\textwidth]{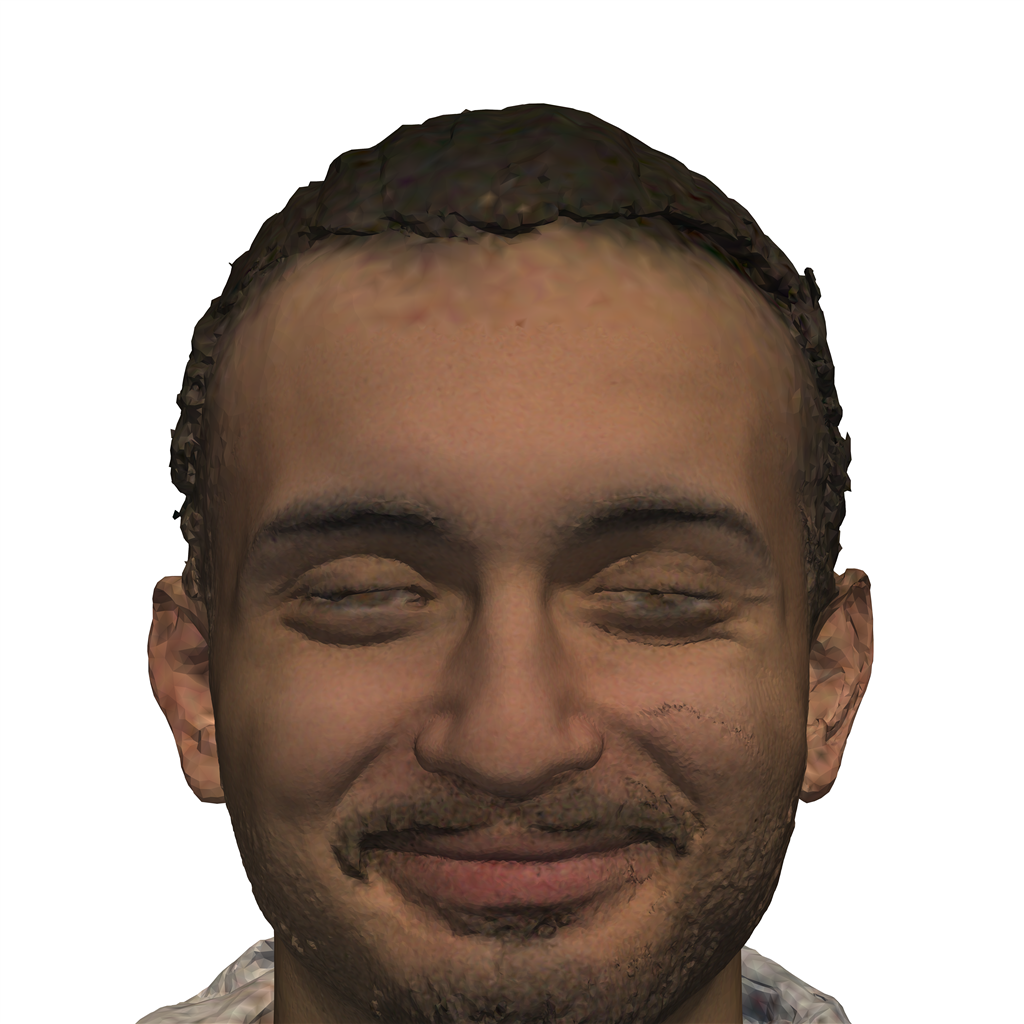} &
    \includegraphics[width=0.3\textwidth]{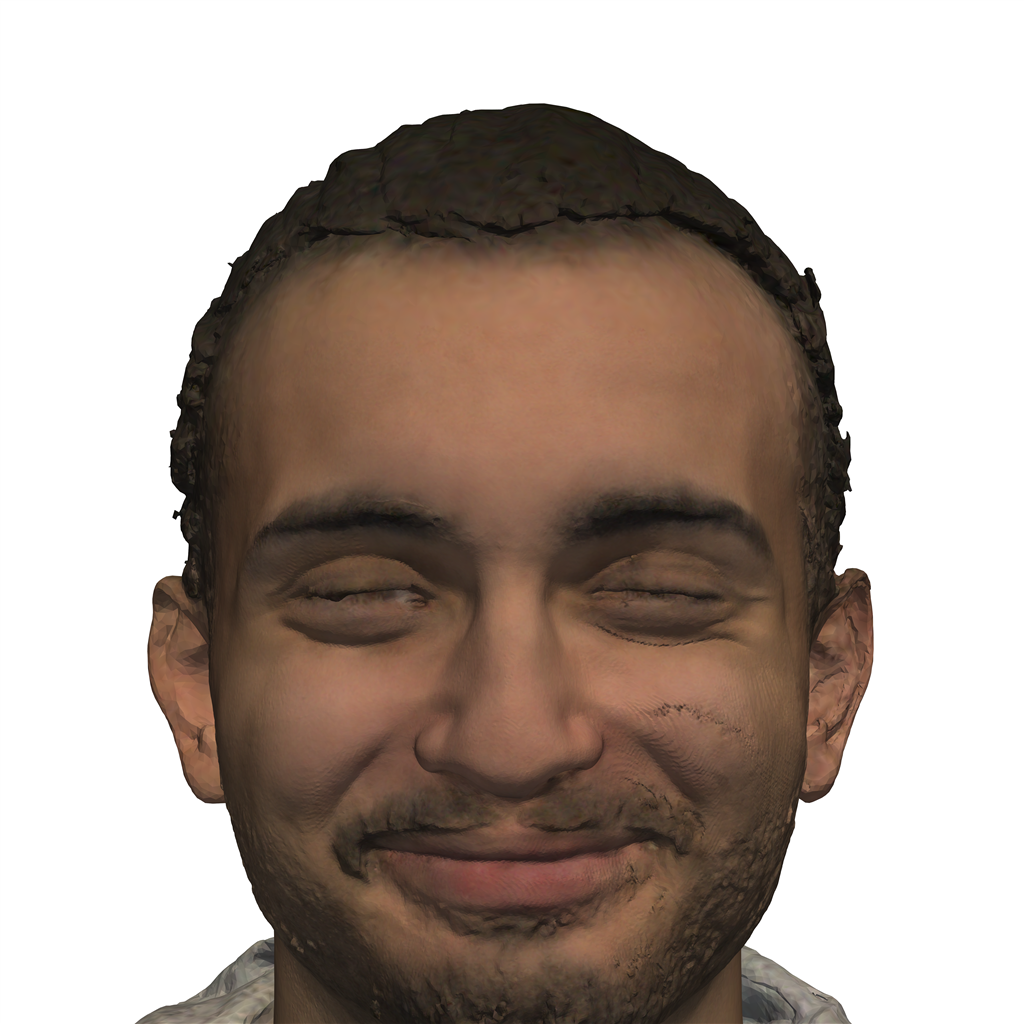} \\

    \includegraphics[width=0.3\textwidth]{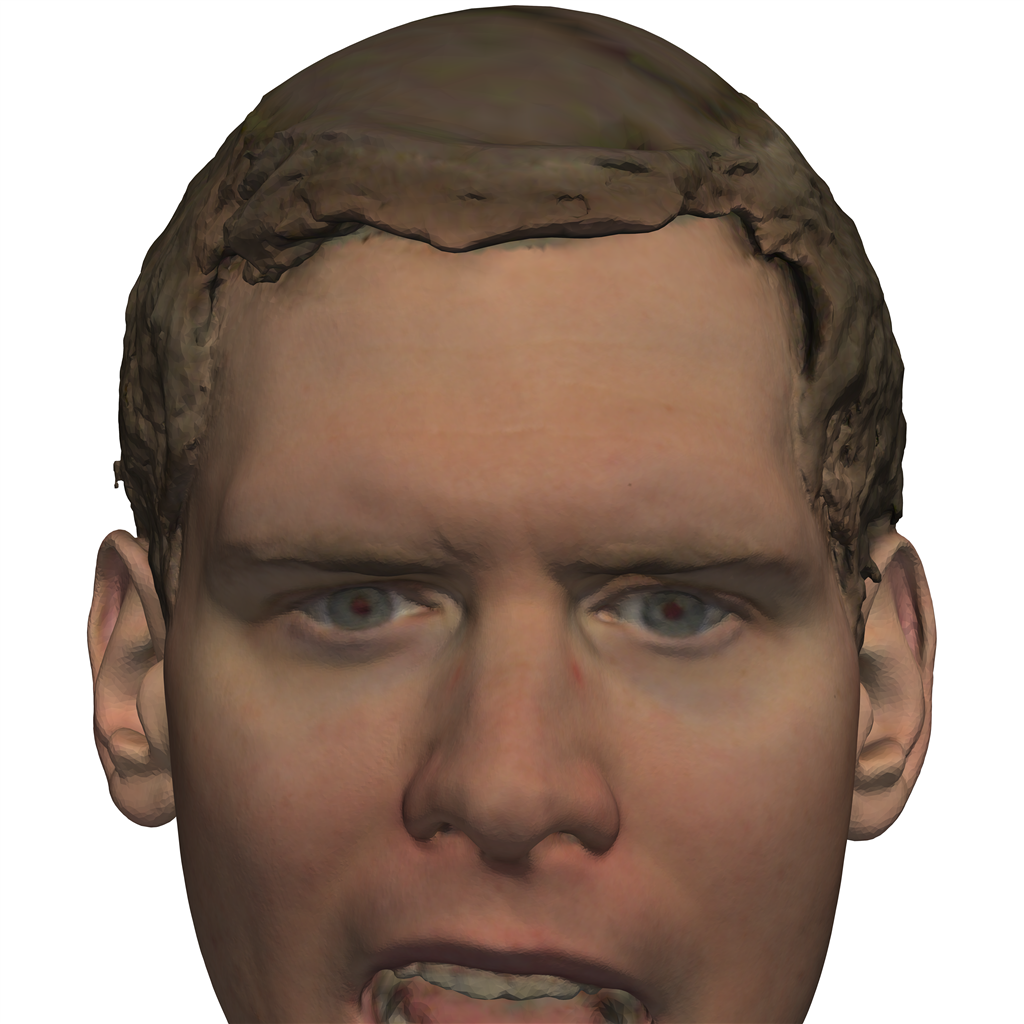} &
    \includegraphics[width=0.3\textwidth]{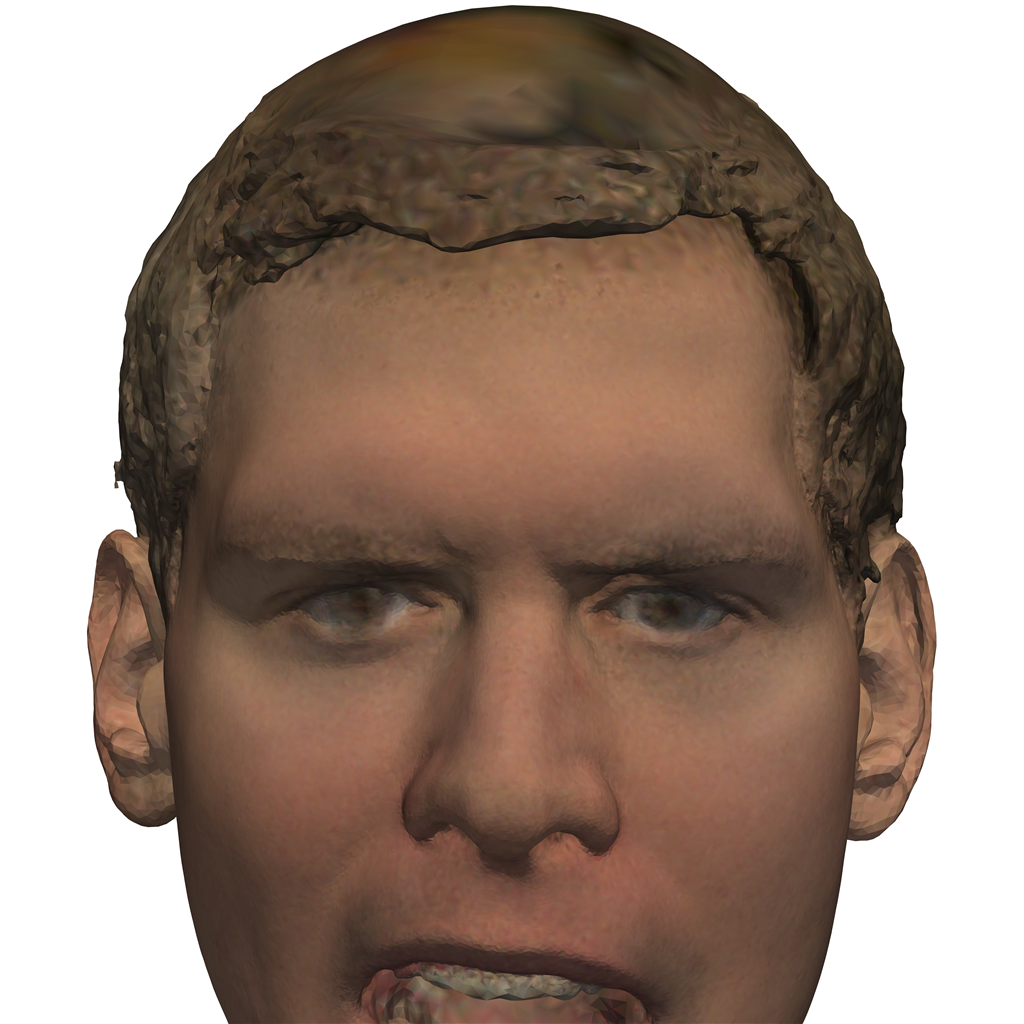} &
    \includegraphics[width=0.3\textwidth]{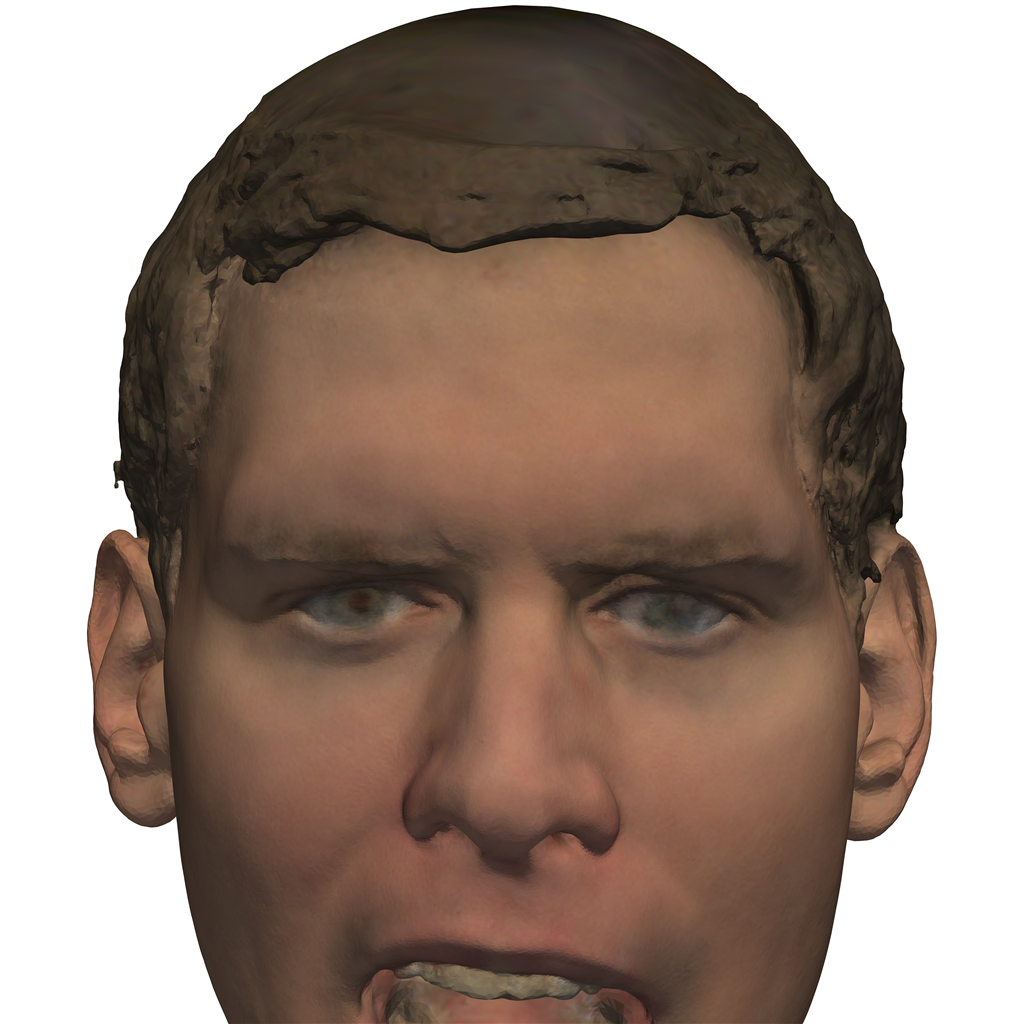} \\

    \includegraphics[width=0.3\textwidth]{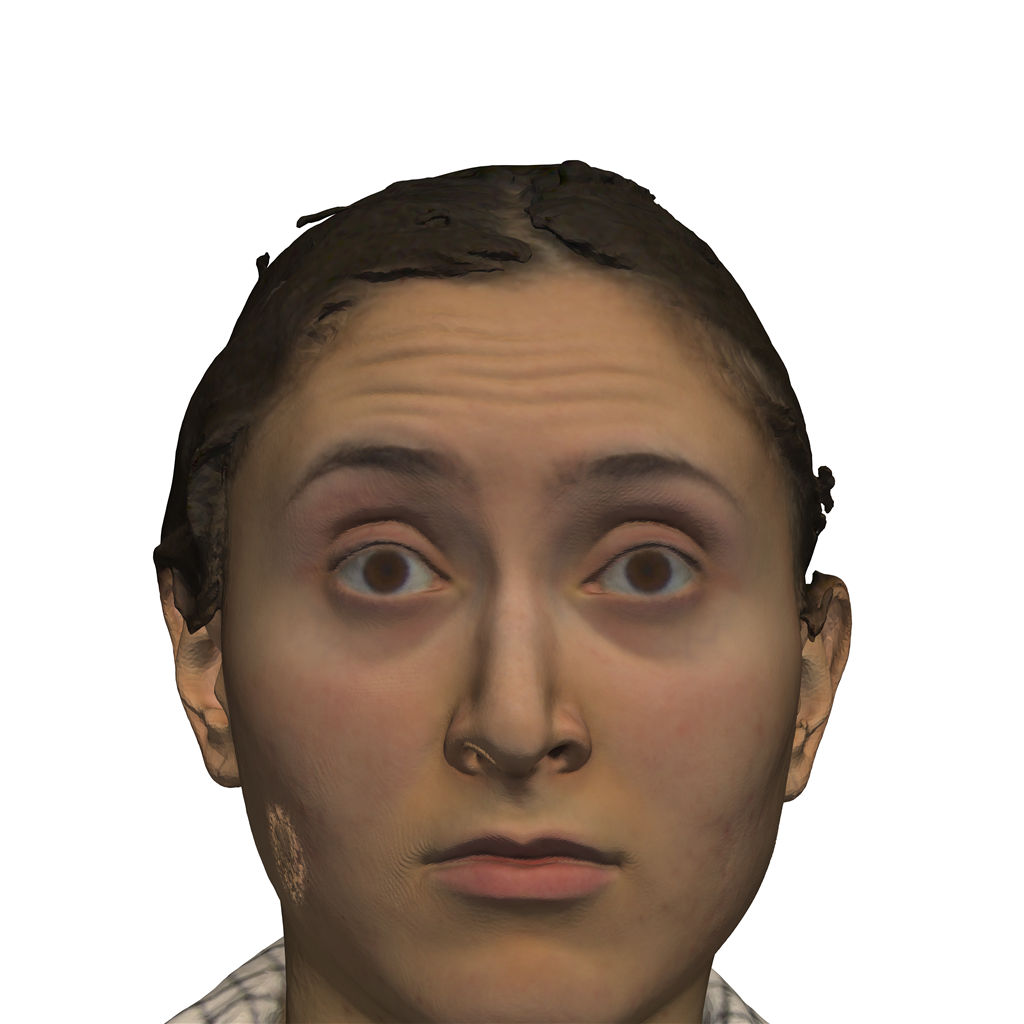} &
    \includegraphics[width=0.3\textwidth]{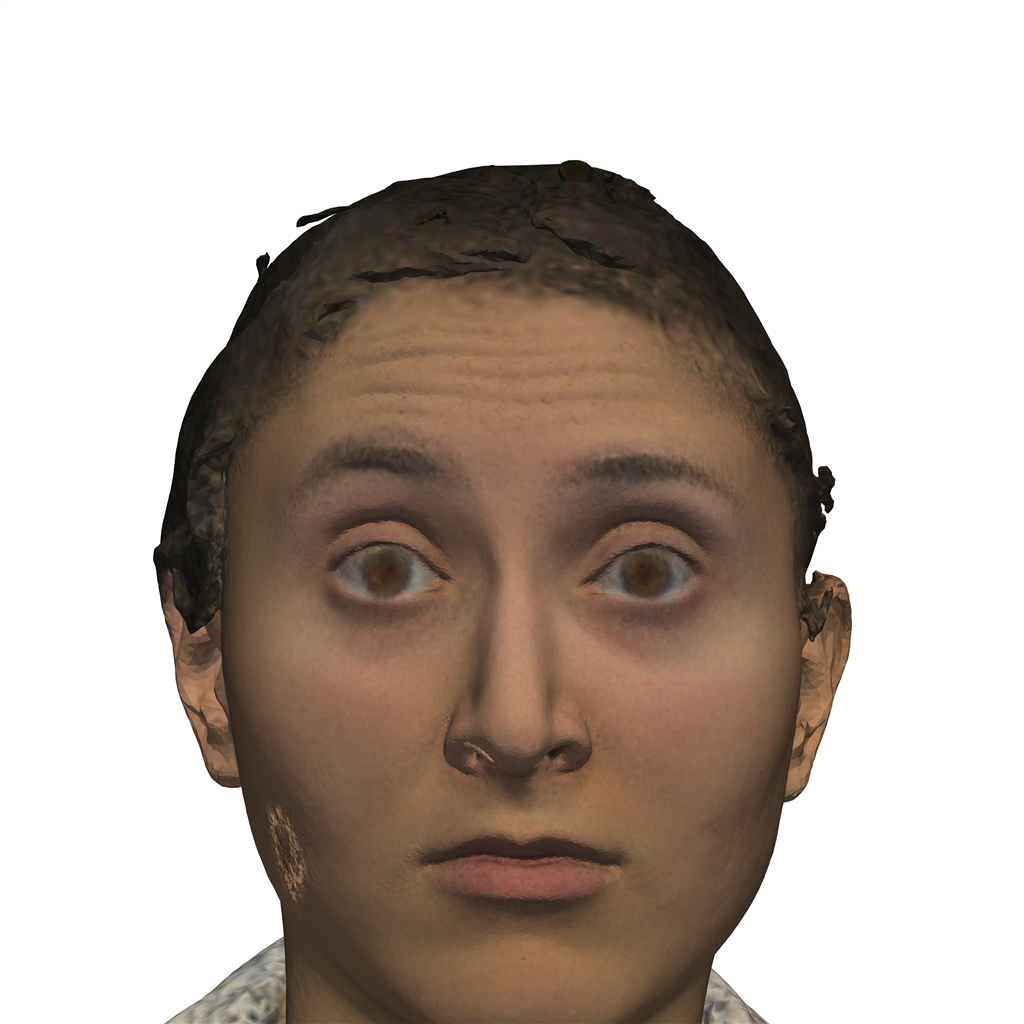} &
    \includegraphics[width=0.3\textwidth]{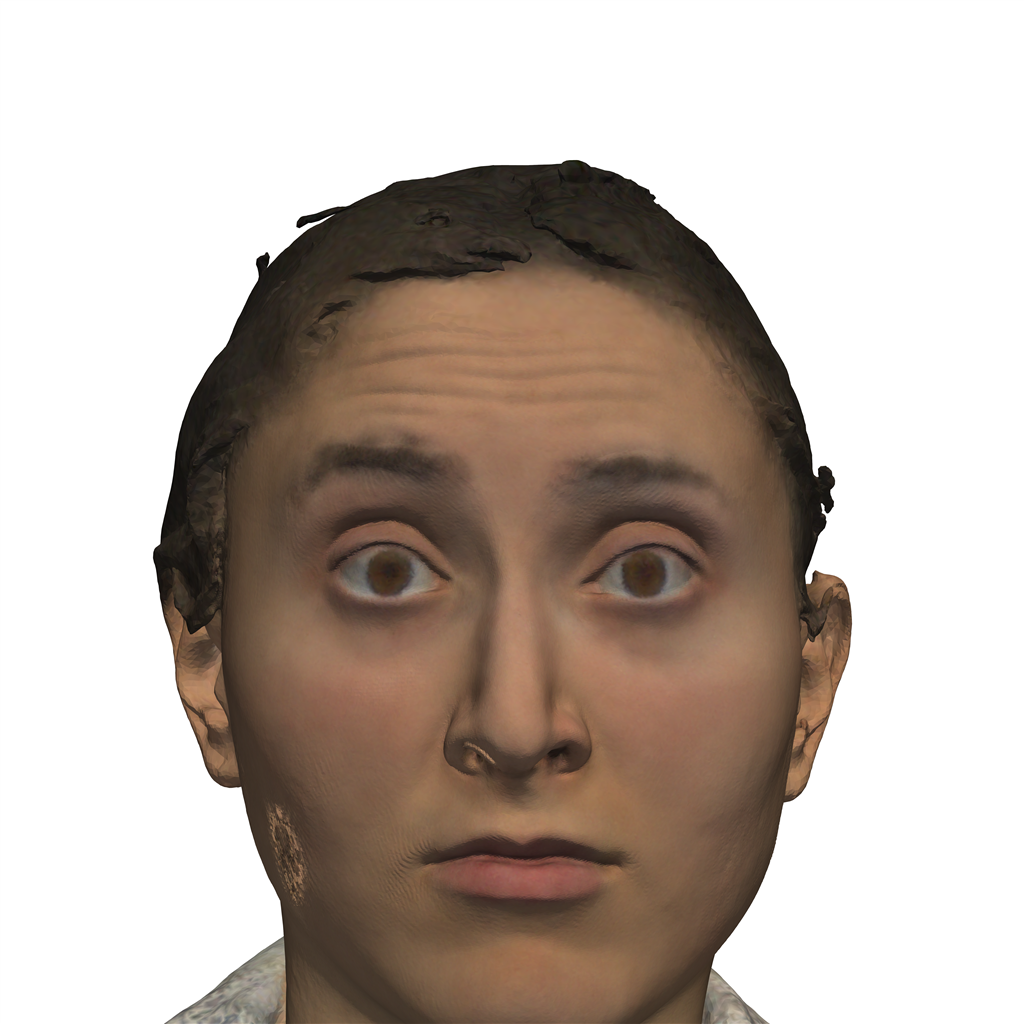} \\
\end{tabular}

\caption{\textbf{Effect of the local high-frequency stream.} Removing
Stream~2 leaves the spectral branch alone, whose truncated
Laplace--Beltrami basis acts as a low-pass filter, and the MLP in each EDB block needs to retain and enhance local features, which leads to a noisy, grainy look.}
\label{fig:supp_hf}
\end{figure*}

Removing the high-frequency stream forces the network to rely solely
on the EDB's per-vertex MLP for detail recovery. As shown in the top
row of \cref{fig:visual_ablation} and in \cref{fig:supp_hf}, this produces a visibly grainy,
locally incoherent skin texture and a notable loss of high-frequency
sharpness in boundaries like hair and skin and in fine structures such as beard stubble.

Disabling noise-level conditioning in the heat diffusion step prevents the network from dynamically adapting the spatial reach of feature propagation. As shown in the middle row, this causes a severe boundary collapse: skin tones bleed into the hair region and dark hair colour
smears down the neck onto clothing --- artefacts that are entirely
absent in the full VISAGE model.

Without semantic region splitting, a shared global MLP is forced to
process all vertex types jointly. As shown in the bottom row, this
degrades the structural integrity of specialised regions: the eyes
lose their iris definition, and the pink hues of the gums bleed
unnaturally into the teeth. The full model maintains strict semantic
boundaries and preserves high-frequency fidelity across all regions.

\subsection{Failure Cases} \label{supp:viz:failures}

We present two representative failure modes to characterise the current
limitations of VISAGE, consistent with the discussion in Sec.~5 of the
main paper.

\begin{figure*}[bpht]
    \centering
    
    \begin{subfigure}[t]{0.22\linewidth}
        \includegraphics[width=\linewidth]{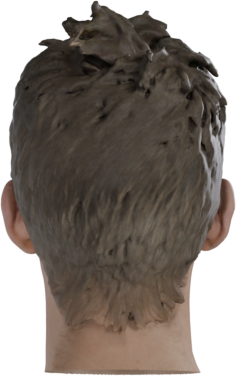}
        \caption*{GT}
    \end{subfigure}\hfill
    \begin{subfigure}[t]{0.22\linewidth}
        \includegraphics[width=\linewidth]{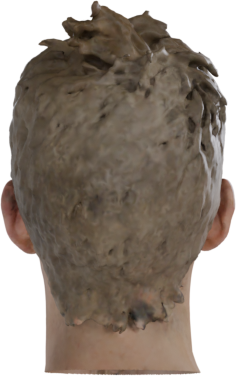}
        \caption*{VISAGE}
    \end{subfigure}\hfill
    \begin{subfigure}[t]{0.24\linewidth}
        \includegraphics[width=\linewidth]{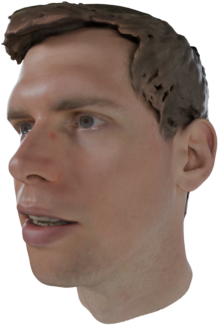}
        \caption*{GT}
    \end{subfigure}\hfill
    \begin{subfigure}[t]{0.24\linewidth}
        \includegraphics[width=\linewidth]{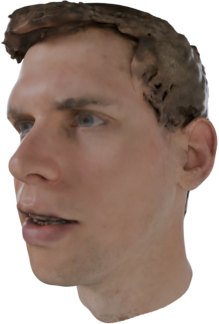}
        \caption*{VISAGE}
    \end{subfigure}
    
    \vspace{0.1cm} 
    \begin{subfigure}[t]{0.24\linewidth}
        \includegraphics[width=\linewidth]{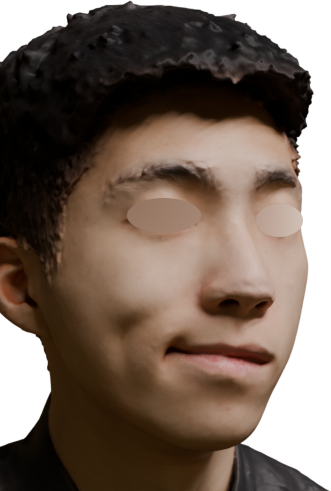}
        \caption*{GT}
    \end{subfigure}\hfill
    \begin{subfigure}[t]{0.24\linewidth}
        \includegraphics[width=\linewidth]{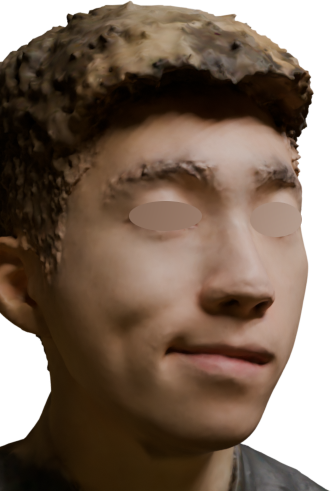}
        \caption*{VISAGE}
    \end{subfigure}\hfill
    \begin{subfigure}[t]{0.22\linewidth}
        \includegraphics[width=\linewidth]{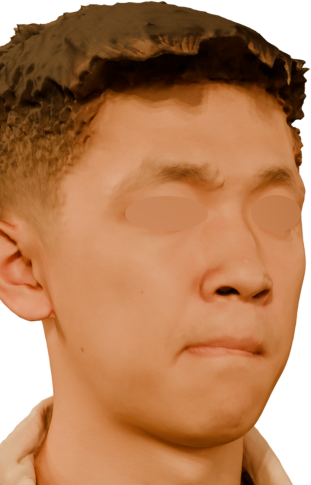}
        \caption*{GT}
    \end{subfigure}\hfill
    \begin{subfigure}[t]{0.22\linewidth}
        \includegraphics[width=\linewidth]{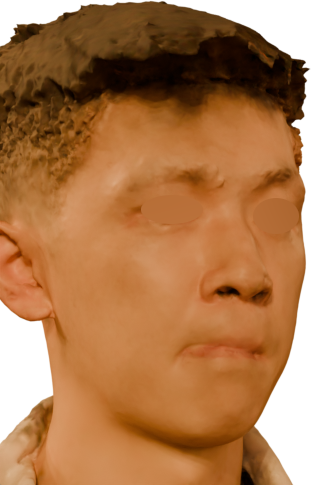}
        \caption*{VISAGE}
    \end{subfigure}

    \caption{Typical failure cases of VISAGE. \textbf{Top Block:} Results from the NPHM dataset \cite{NPHM}, where sometimes the colour in the occluded hair region shifts slightly and gets brighter, compared to hair colour in no occluded areas on the right. \textbf{Bottom Block:} Results from FaceVerse dataset \cite{FaceVerse}, in which strong directional lighting and harsh cast shadows are occasionally baked into the predicted albedo rather than being fully decomposed. }
    \label{fig:failure_cases}
\end{figure*}

\paragraph{Occluded Regions.}
Since direct image supervision is limited to frontal and lateral views,
texture synthesis in heavily occluded regions --- primarily the hair at the back of the head --- is governed entirely by the diffusion prior.
As shown in the top block of \cref{fig:failure_cases}, this occasionally
produces a subtle luminance shift, where the predicted hair colour
becomes slightly brighter than the ground truth.
This artefact is most
noticeable under full $360^\circ$ inspection.

\paragraph{Illumination Entanglement.}
Under strong directional lighting or harsh cast shadows, the network occasionally fails to fully disentangle surface albedo from incident
illumination, noticeable most in the hair region. As shown in the bottom block of \cref{fig:failure_cases},
this manifests as a warm colour cast baked into the predicted albedo,
most visibly in dark hair and shadowed skin regions. Sec.~4.2.3 of the main paper quantifies
the effect across ten environments: predicted albedo removes $83\%$ of the
observed illumination variation, with the residual concentrated in hair.
Recovering the remainder would require explicit illumination decomposition
rather than the single-pass prediction used here.
\clearpage
\clearpage

\end{document}